\PassOptionsToPackage{table,xcdraw,dvipsnames}{xcolor}

\documentclass[11pt,letterpaper]{plum}

\usepackage[all]{hypcap}
\PassOptionsToPackage{round}{natbib}
\usepackage{natbib}
\usepackage{pifont}
\usepackage{nicefrac}
\usepackage{arydshln}
\usepackage{subcaption}
\usepackage{pgfplots}
\pgfplotsset{compat=1.18}
\usepgfplotslibrary{groupplots}
\tcbuselibrary{most,skins,theorems}
\usepackage{listings}
\usepackage{xspace}
\usepackage{soul}
\usepackage{cleveref}
\usepackage{etoc}
\usepackage[section]{placeins}

\definecolor{darkblue}{rgb}{0,0,0.5}
\definecolor{lightblue}{RGB}{220,235,250}
\definecolor{tabhead}{HTML}{DCE6F1}
\definecolor{tabband}{HTML}{F2F2F2}
\definecolor{cellgood}{HTML}{D8ECD2}
\definecolor{cellbad}{HTML}{F4D2D2}
\definecolor{cellbase}{HTML}{E8E8E8}
\definecolor{textgood}{HTML}{1B7A2D}
\definecolor{textbad}{HTML}{B22222}
\definecolor{textmute}{HTML}{6B6B6B}

\hypersetup{colorlinks=true,citecolor=darkblue,linkcolor=darkblue,urlcolor=darkblue}

\definecolor{lightorange}{HTML}{faa755}
\definecolor{annothl}{RGB}{255,232,150}
\definecolor{annotaccent}{RGB}{170,40,30}
\definecolor{diffaccent}{RGB}{30,90,170}
\sethlcolor{annothl}

\tcbset{
  takeawaysbox/.style={
    title=Takeaway,
    colback=lightblue!80,
    colframe=black,
    fonttitle=\bfseries\small,
    coltitle=white,
    colbacktitle=black,
    enhanced,
    attach boxed title to top left={xshift=2.5mm,yshift=-2.5mm},
    boxed title style={rounded corners, size=small, colframe=black, colback=black},
    width=\linewidth,
    arc=3.5mm
  },
  taskbox/.style={
    title=Task,
    colback=lightorange!10,
    colframe=lightorange!80!black,
    fonttitle=\bfseries\small,
    coltitle=white,
    colbacktitle=lightorange!80!black,
    enhanced,
    attach boxed title to top left={xshift=2.5mm,yshift=-2.5mm},
    boxed title style={rounded corners, size=small, colframe=lightorange!80!black, colback=lightorange!80!black},
    width=\linewidth,
    arc=2mm
  },
  promptbox/.style={
    colback=lightorange!10,
    colframe=black,
    fonttitle=\bfseries\small,
    coltitle=white,
    colbacktitle=black,
    enhanced,
    breakable,
    attach boxed title to top left={xshift=2.5mm,yshift=-2.5mm},
    boxed title style={rounded corners, size=small, colframe=black, colback=black},
    width=\linewidth,
    arc=2mm
  },
  examplebox/.style={
    colback=gray!5,
    colframe=gray!50,
    boxrule=0.5pt,
    arc=2mm,
    width=\linewidth,
    fontupper=\small\ttfamily,
    enhanced,
    breakable,
    left=2mm,right=2mm,top=1.5mm,bottom=1.5mm
  }
}

\lstdefinestyle{solverpy}{
  basicstyle=\ttfamily\scriptsize,
  language=Python,
  breaklines=true,
  columns=fullflexible,
  keepspaces=true,
  showstringspaces=false,
  aboveskip=0pt, belowskip=0pt,
  commentstyle=\color{gray!60!black}\itshape,
  keywordstyle=\color{blue!60!black}\bfseries,
  stringstyle=\color{green!40!black}
}

\newcommand{\projectname}{\emph{CausaLab}}

\title{\projectname: A Scalable Environment for Interactive Causal Discovery Toward AI Scientists}

\author[1]{Junlin Yang\textsuperscript{*}}
\affil[1]{Tsinghua University}
\author[2]{Dylan Zhang\textsuperscript{*}}
\affil[2]{University of Illinois Urbana-Champaign}
\author[3]{Xiangchen Song}
\affil[3]{Carnegie Mellon University}
\author[4]{Qirun Dai}
\affil[4]{University of Chicago}
\author[4]{Xiao Liu}
\author[2]{Yuen Chen}
\author[2]{Aniket Vashishtha}
\author[5]{Jing Shi}
\affil[5]{Adobe}
\author[4]{Chenhao Tan}
\author[2]{Hao Peng}
\correspondingauthor{Dylan Zhang, \href{mailto:shizhuo2@illinois.edu}{shizhuo2@illinois.edu}}

\begin{abstract}
We introduce \projectname{}, a scalable environment for evaluating interactive causal discovery by LLM agents. Unlike prior evaluations, \projectname{} evaluates both whether an agent can solve a problem using causal evidence and whether its answer is grounded in a faithful recovered causal mechanism. Each episode places an agent in a synthetic laboratory: it receives prior measurement records, intervenes on a manipulator crystal, and predicts the resonance frequency of a held-out reactor crystal governed by the same mechanism. The hidden data-generating process is a randomly sampled structural causal model (SCM), so success requires recovering both a causal graph and structural equations rather than recalling prior knowledge.
Experiments show a persistent gap between prediction and mechanism recovery: in the purely observational 6-node setting, \texttt{GPT-5.2-high} reaches 92\% task accuracy but only 0.471 all-edge $F_1$.
Mixed observation--intervention strategies improve structural fidelity, while pure intervention remains difficult even for strong agents.
We identify premature stopping as a major weakness and show that consistency verification mitigates it.
\projectname{} therefore separates predictive success from causal understanding and exposes current LLM agents' limits as experimental causal reasoners.

\noindent{\footnotesize Code: \url{https://github.com/DylanZSZ/CausaLab}}

\noindent{\footnotesize\textsuperscript{*}Junlin Yang and Dylan Zhang contributed equally and both serve as project leads. Junlin Yang's work was done at the University of Illinois Urbana-Champaign.}
\end{abstract}

\begin{document}

\maketitle
\makeatletter
\@thanks
\let\@thanks\@empty
\makeatother

\section{Introduction}
Causal reasoning is important because scientific, medical, and policy decisions depend on how systems would respond to interventions, not only on observed associations \citep{Pearl_2009,pearl2018why,imbens_rubin_causal_2015}.
Yet measuring and making progress in causal reasoning remains challenging, particularly for today's large language models (LLMs).
Existing benchmarks generally
 translate causal graphs, datasets, or narratives into question-answering and classification tasks \citep{qin2019counterfactual,romanou2023crab,stolfo2023causal,jiang2024can,vashishtha2025teachingtransformerscausalreasoning,jin2024cladderassessingcausalreasoning,wang-2024-causalbench,clear2024,corr2cause2024}.
While useful, they leave open the ``causal parrot'' concern \citep{zecevic2023causal}:
models can succeed with memorized causal facts or linguistic cues rather than causal reasoning behaviors needed to discover causal mechanisms \citep{zheng2023preservingcommonsenseknowledgepretrained,liu2023kept}.

\begin{figure*}[t]
\centering
\includegraphics[width=1.0\linewidth]{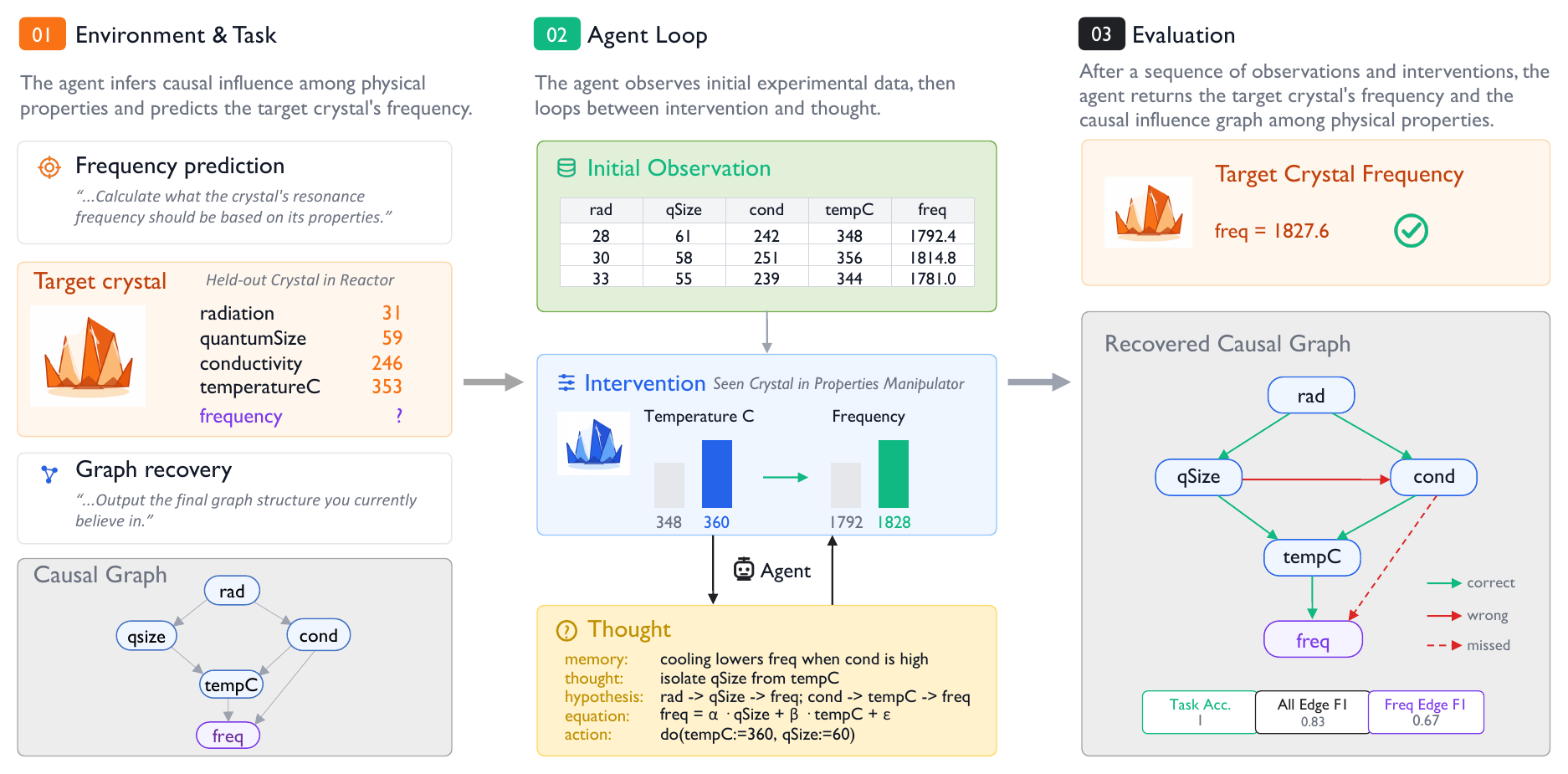}
\caption{Overview of a \projectname{} episode. \textbf{(1)} A hidden SCM generates prior records, a manipulator crystal, and a held-out reactor crystal. \textbf{(2)} The agent observes records and performs budgeted interventions on the manipulator crystal. \textbf{(3)} At each step it emits a DSL thought parsed against the ground-truth SCM. \textbf{(4)} It predicts the reactor \texttt{frequency}; we score both the prediction and recovered-mechanism trajectory.}
\label{fig:causal_main}
\end{figure*}

To illustrate, let's consider the following thought experiment. Suppose we are interested in studying the causal relationship between temperature and the resonance frequency of a crystal. An LLM agent might appear useful in at least two different ways. (1) It may retrieve from existing sources, such as Wikipedia or its training data, that temperature causes resonance frequency.
(2) It may observe paired measurements of temperature and frequency, formulate hypotheses, design experiments, perform interventions, observe the resulting changes, and infer causation from evidence \citep{Pearl_2009,JMLR:v13:hauser12a,lampinen2023passive}. While both are valuable in practice, (1) offers little help when the relevant causal knowledge lies beyond the current frontier of human knowledge.
We therefore argue that (2) is especially important, particularly for important applications such as scientific discovery, because it enables LLM agents to help advance the frontiers of knowledge in a manner closer to what human scientists would do \citep{langley2019scientific,dunbar_fugelsang_2005,DBLP:journals/corr/abs-2406-06769}.

We introduce \projectname{} (Figure~\ref{fig:causal_main}), a scalable environment for evaluating LLM agents as interactive causal discoverers, joining a recent line of interactive scientific-agent and causal-discovery benchmarks \citep{DBLP:journals/corr/abs-2406-06769,igda2025,chen2026causalgame,chen2025autobench,geng2025reliableaiscientists}. Each episode asks the agent to use evidence from prior records and interventions on one crystal to predict the held-out frequency of another crystal. The shared data-generating mechanism is a hidden structural causal model (SCM) \citep{Pearl_2009}, with a causal graph and structural equations that determine the crystal properties and frequency.
The agent receives prior measurement records, can run budgeted interventions on a manipulator crystal through a property manipulator, and must predict the frequency of a separate reactor crystal governed by the same SCM (Figure~\ref{fig:causal_main}; \S\ref{sec:benchmark}). Two design choices distinguish \projectname{} from prior causal-reasoning evaluations. First, the hidden SCM is sampled per episode rather than drawn from public causal corpora, which sidesteps the ``causal parrot'' concern that scores reflect memorized causal lexicon. Second, a lightweight domain-specific language (DSL; \S\ref{sec:dsl}) records the agent's accumulated evidence, current graph and equation hypothesis, planned experiment, and action at each step, so we can score not only the final prediction but also the trajectory-level faithfulness of the recovered mechanism to the ground-truth SCM (\S\ref{sec:experiments}).

Our experiments span closed and open-weight models, multiple model sizes, and thinking versus non-thinking variants, surfacing four findings that prior static benchmarks cannot reach. \textbf{(1) Correct predictions often do not reflect correct mechanism discovery.} Across matched functional-form controls, hidden-perturbation controls, and target-edge controls, endpoint accuracy and mechanism fidelity move separately: agents can find plausible parents while missing quantitative equations, preserve task success while degrading all-edge recovery, or lose accuracy mainly when the target equation itself is perturbed. \textbf{(2) Observation-conditioned online intervention best balances prediction and graph recovery.} Pure observation can boost endpoint accuracy without recovering structure, and pure intervention is weak before observations narrow the hypothesis space. For \texttt{GPT-5.2-high} on 6-node graphs, pure observation reaches 92\% accuracy but only 0.47 all-edge $F_1$, while mixed online observation--intervention reaches 80\%/0.80. Offline intervention traces do not replace online experimental choice: injecting ``Golden'' chains raises \texttt{GPT-5-mini} accuracy to 90\% on 4 nodes while \emph{lowering} all-edge $F_1$. \textbf{(3) Model family and scale pay off unevenly across the two axes.} \texttt{GPT-5.2-high} has the best endpoint accuracy and lowest directed all-edge structural Hamming distance (SHD) at every graph size, but gains are not uniform across graph sizes or metrics. Open-weight \texttt{Qwen3.5} can approach \texttt{GPT-5-mini} on some task scores, yet its SHD rises faster as graphs grow; thinking generally lowers Qwen SHD. Even \texttt{GPT-5.2-high} drops to 64\% accuracy and directed SHD 4.761 at 7 nodes. \textbf{(4) Many failures come from premature commitment, not exhausted budget.} Both successful and failed runs leave roughly half the intervention budget unspent, failed runs end with hypotheses inconsistent with their own data, and a single explicit verification step lifts 4-node accuracy from 48\% to 60\%. \projectname{} therefore separates predictive success from causal understanding, revealing how current LLM agents still struggle to explore unfamiliar environments interactively, test candidate mechanisms, and revise toward the causal regularities that govern them.

\section{Background and Related Work}

Causal reasoning goes beyond associational prediction by asking how a system would change under interventions and counterfactual alternatives \citep{Pearl_2009,pearl2018why,imbens_rubin_causal_2015}. Structural causal models (SCMs) formalize these assumptions as directed graphs plus structural equations \citep{Pearl_2009}. In \projectname{}, each episode's hidden SCM is both the ground truth (\S\ref{sec:scm-construction}) and the evaluation target (\S\ref{sec:eval-targets}), letting us score whether an agent recovers the graph and target equation, not only whether it predicts the reactor value.

Most LLM causal evaluations are static: they ask models to answer textual causal questions, reason over given graphs, classify cause--effect direction, or solve formal causal-inference queries \citep{kiciman2023causal,jin2024cladderassessingcausalreasoning,corr2cause2024,clear2024,wang-2024-causalbench,chen2024causal}. Related work also uses LLMs as causal priors for edge scoring, causal ordering, or query-efficient discovery \citep{llmcausalgraphs2023,llmeffectivepriors2024,causalorder2023,efficientcausalgraph2024}. Recent SCM-oriented studies either use LLM metadata reasoning to support graph discovery \citep{abdulaal2024causalmodellingagents} or test coefficient elicitation when the DAG is supplied \citep{yamaoka2026linearllmscm}. HypoBench further shows that hypothesis-generation benchmarks must account for how prior knowledge shapes model behavior \citep{liu2025hypobench}. These settings clarify what causal knowledge LLMs can express, but they usually provide the variables, evidence, graph, or query up front. \projectname{} instead asks whether an LLM agent can gather evidence, revise a hypothesis, and transfer the learned mechanism to a new instance, all within a scientific-discovery setting that offers no hints about the underlying causal structure.

Interactive environments broaden evaluation beyond one-shot answers, including scientific-discovery worlds, budgeted graph-discovery games, causal games, and non-LLM intervention planners \citep{DBLP:journals/corr/abs-2406-06769,igda2025,chen2026causalgame,gregorini2025dodo}. A basic agent scaffold for such settings is ReAct-style reasoning and acting, where the model interleaves deliberation with executable environment actions \citep{DBLP:conf/iclr/YaoZYDSN023}. The closest recent benchmark is Auto-Bench, where LLM agents iteratively query scientific or social-network environments to recover a hidden causal graph \citep{chen2025autobench}. Work on black-box reverse engineering similarly shows that actively designing queries is not equivalent to receiving another agent's intervention data \citep{geng2025reliableaiscientists}.

\projectname{} differs from Auto-Bench in its evaluation target. Auto-Bench primarily asks whether an agent can discover a hidden DAG through interaction. \projectname{} asks whether the discovered mechanism \emph{transfers}: after learning from prior measurements and interventions on a manipulator crystal, the agent must predict a held-out reactor crystal generated by the same SCM, while its per-step DSL hypotheses expose the graph, the \texttt{frequency} structural equation, and the coefficients it is committing to. This makes it possible to separate task utility from structural and quantitative faithfulness, and to audit how an LLM agent revises or fails to revise an explicit SCM over time.

This connects two evaluation practices: explicit SCM recovery from causal discovery and sequential experiment design from agent benchmarks. Because each episode has a known ground-truth SCM and a logged interaction trace, \projectname{} can score both final-task utility and the faithfulness of the recovered mechanism.

\section{The Construction of CausaLab}
\label{sec:benchmark}

This section first defines the episode-level task and what the agent must
infer, then specifies the SCM in \S\ref{sec:scm-construction}, the observation
and intervention protocol in \S\ref{sec:obs-int}, and the evaluation targets in
\S\ref{sec:eval-targets}. Artifact, licensing, and implementation details
are provided in Appendix~\ref{app:artifact_implementation}.
Throughout the section, Figure~\ref{fig:causal_main} serves as a running
example: the agent first observes prior crystal records, then intervenes on a
controllable property of the manipulator crystal, and finally predicts the
reactor crystal's hidden \texttt{frequency}.

\paragraph{Design principles.}
The benchmark is designed around three goals. First, can a model infer a
causal mechanism that transfers to a new instance, rather than fitting an
isolated value pattern? Second, can it choose informative interventions rather
than passively consume a fixed dataset? Third, how do these abilities scale
with graph size, topology, functional form, intervention budget, and hidden
disturbances? The corresponding design choices that realize these goals are
shared-mechanism transfer between two crystals, online intervention choice,
and synthetically controlled SCM generation with known ground truth.

\paragraph{Task formulation.}
A \projectname{} episode is a transfer problem under a hidden SCM: the causal
graph, structural equations, and coefficients are all hidden, and the agent is
given only prior measurement records plus a finite budget for interventions
(Figure~\ref{fig:causal_main}). The episode also contains two crystals
generated by the same SCM: a manipulator crystal on which the agent may
intervene, and a reactor crystal whose \texttt{frequency} is held out. The
initial records contain physical properties and resulting \texttt{frequency}
values from earlier measurements under the same SCM. The agent then spends its interaction budget on
interventions over controllable non-\texttt{frequency} properties of the
manipulator crystal and observes the resulting measurements. After collecting
this evidence, the agent predicts the hidden \texttt{frequency} of the reactor
crystal. The records, manipulator crystal, and reactor crystal share the same
SCM but have different property values, so the agent cannot solve the task by
copying an observed frequency; it must infer a mechanism that transfers.

The agent is told the property names and functional family but receives
interventions only on a configured subset \(C\subseteq O\) of controllable
observable non-\texttt{frequency} variables; variables outside \(C\)
(including \(Y\) and any non-controllable property) are observable but not
intervenable. The reactor crystal exposes only its non-\texttt{frequency}
variables; Appendix Table~\ref{tab:variable_access} summarizes which variables
are observable, intervenable, and hidden/exogenous in the episode. At each step
the agent also emits a DSL
hypothesis that we parse into a directed graph, \texttt{frequency} equation,
and coefficients. Solving an episode therefore requires both a correct
reactor prediction and a causal hypothesis that matches the hidden SCM under
the metrics of \S\ref{sec:eval-targets}.

\subsection{Structural Causal Models}
\label{sec:scm-construction}

Each episode instantiates an SCM
\(\mathcal{M}=(\mathbf{U},\mathbf{V},F,P(\mathbf{U}))\) \citep{Pearl_2009}.
Here \(\mathbf{U}\) are exogenous source terms, \(\mathbf{V}\) are endogenous
variables, \(F\) is the set of structural equations, and \(P(\mathbf{U})\) is
the exogenous distribution. In \projectname{}, the endogenous variables are
observable properties \(O\) plus the target \(Y=\texttt{frequency}\). Root
variables are endogenous nodes whose values are generated from exogenous source
terms, and optional hidden-noise terms are also exogenous. We sample a DAG
\(G\) over \(\mathbf{V}=O\cup\{Y\}\), assign root nodes from their exogenous
sources, then compute non-root variables in topological order. We use exactly
two structural-equation families: linear and quadratic. In the linear family,
\[
X = b + \sum_{p \in \mathrm{pa}(X)} w_p p,
\]
and in the quadratic family,
\[
X = b + \sum_{p \in \mathrm{pa}(X)} (u_p p^2 + w_p p).
\]
The sampled graph, equations, and coefficients, including the base value of
\texttt{frequency}, are shared across the prior records, manipulator crystal,
and reactor crystal; controllable-property base values differ across these
instances. This hidden
SCM corresponds to the causal graph in Figure~\ref{fig:causal_main}, serves as the common mechanism behind the prior records, the manipulator crystal,
and the reactor crystal. This asymmetry is what forces the agent to infer how variables are
connected and then apply that mechanism to the reactor's property values.

Some graph families also include an unobserved exogenous disturbance \(H\)
that perturbs the system as follows. After every intervention, \(H\) is
resampled and added as a fixed-weight shift to a designated subset of
observable endogenous variables; those shifted values then propagate
downstream through \(F\). \(H\) itself is not in \(\mathbf{V}\), is not named
to the agent, and cannot be observed or set directly --- the agent sees only
its downstream effects on the returned variable values. These settings test
whether an agent can distinguish a stable causal mechanism from
post-intervention noise. Additional distributions and coefficient ranges
appear in Appendix~\ref{app:appendix}; formal SCM and hidden-disturbance
details appear in Appendix~\ref{app:scm_details}.
\subsection{Interaction and Outputs}
\label{sec:obs-int}

Each episode proceeds through a repeated hypothesis--experiment loop. The
agent receives an initial batch of measurement records, including
non-\texttt{frequency} properties and the resulting \texttt{frequency}. It may
then intervene by setting one controllable non-\texttt{frequency} property on
the manipulator crystal; the environment recomputes that crystal's resulting
measurement under the hidden SCM and returns it to the agent. The reactor
crystal is observed but not intervened on: its non-\texttt{frequency}
properties are visible, and its
\texttt{frequency} remains hidden until the agent submits a final value.

Concretely, the loop begins with the initial observation batch and then
alternates between interventions and observations:
\emph{choose an intervention on one controllable manipulator-crystal property}
\(\rightarrow\) \emph{observe the resulting manipulator-crystal measurement}
\(\rightarrow\) \emph{revise the DSL hypothesis and choose the next
intervention}. For example, after seeing several prior measurement records, an
agent may set the manipulator crystal's \texttt{radiation} to a
chosen value, see how \texttt{temperature}, \texttt{conductivity}, and
\texttt{frequency} change, and then decide whether the evidence supports a
direct edge into \texttt{frequency} or an indirect path through another
property. This is the interaction that Figure~\ref{fig:causal_main} depicts at
the task level and Appendix Figure~\ref{fig:causal_visualization} exposes at the
trajectory level.

The intervention semantics are shift-style rather than hard
\(\mathrm{do}(X{=}v)\) \citep{backshift2015}, and we specify them here because
they determine what the agent's returned observations mean.
This models a laboratory control that shifts a controllable baseline while preserving upstream dependencies across sequential interventions.

For a controllable
variable \(X\in C\), an intervention request with value \(v\) replaces the
base term in that variable's structural equation for the next environment
update:
\[
X \leftarrow v + \sum_{p \in \mathrm{pa}(X)} w_p p
\]
in the linear family, and analogously
\[
X \leftarrow v + \sum_{p \in \mathrm{pa}(X)} (u_p p^2 + w_p p)
\]
in the quadratic family. Incoming parent contributions are therefore retained;
only the intercept/base component is shifted. A hard intervention would instead
force \(X=v\) and sever incoming causal influence.

At the end of the episode, the agent submits a numeric prediction for the
reactor \texttt{frequency} and a final DSL hypothesis specifying causal edges,
the proposed structural equation for \texttt{frequency}, and coefficients. The
same DSL can be emitted at intermediate steps, giving a trajectory of evolving
hypotheses.

\subsection{Evaluation}
\label{sec:eval-targets}

We evaluate whether the model both solves the held-out task and recovers the
mechanism needed to solve it causally. Task success is frequency accuracy on
the reactor crystal, corresponding to the final reactor prediction in
Figure~\ref{fig:causal_main}.
Mechanism recovery compares the parsed structured hypothesis log
against the ground-truth SCM: graph precision, recall, and \(F_1\) measure
recovered causal edges; structural Hamming distance (SHD) counts missing,
extra, and reversed directed edges, with lower values indicating closer graph
recovery; coefficient \(F_1\) measures whether the quantitative
\texttt{frequency} mechanism is correct; and root-node identification measures
whether the agent distinguishes exogenous/root variables from mediated
variables. This separation is essential: an agent may predict the held-out
\texttt{frequency} without recovering the SCM, or recover the qualitative graph
while missing the coefficients needed for reliable transfer.
A correct solution therefore requires three linked behaviors: collect useful
observational/interventional evidence, infer a graph and target equation that
explain the prior records and manipulator-crystal measurements, and apply that mechanism to the reactor
crystal's observed properties.

\section{A DSL for Causal Trajectories}
\label{sec:dsl}

Final-answer accuracy cannot distinguish guessing from transferable mechanism discovery. We therefore introduce a domain-specific language (DSL) that records per-step causal commitments and converts hypotheses into SCM artifacts for trajectory-level scoring.

At each interaction step \(t\), the agent emits a compact DSL record with five
fields: \emph{Memory} \(M_t\), the persistent episode notes; \emph{Thought}
\(T_t\), a short interpretation of the current evidence; \emph{Past data}
\(\mathcal{D}_{\le t}\), the accumulated observations and intervention
outcomes; \emph{Hypothesis} \(H_t\), the current causal claim; and
\emph{Experiment} \(E_t\), the next planned intervention and its rationale.
Only \(H_t\) is used as a scored causal artifact: it states hypothesized edges,
the structural equation for \texttt{frequency}, and the associated
coefficients. Appendix Figure~\ref{fig:causal_visualization} shows how parsed
hypotheses are rendered as candidate graphs and recovery metrics over time.
Prompting and repair details appear in Appendix~\ref{app:dsl_implementation}.

\paragraph{Making the hypothesis parsable.}
We make \(H_t\) a scored object by requiring a fixed schema rather than
free-form prose. The schema contains three typed parts: directed edges as
\texttt{(parent, child)} pairs over episode variables, a \texttt{frequency}
structural equation in the declared functional family, and numeric coefficients
for the equation terms. A deterministic parser converts each valid hypothesis
into a candidate graph \(G_t\) and target mechanism \(\hat f_t\), producing a
trajectory \(\{(G_t,\hat f_t)\}_{t=1}^{T}\). This lets the benchmark score the
mechanism the agent commits to at each step using the same graph, root, and
coefficient metrics used for final evaluation, rather than relying only on the
final numeric answer.

\section{Experiments}
\label{sec:experiments}

We use \projectname{} to ask four questions. (RQ1) Does correct prediction
imply mechanism recovery? (RQ2) Which interaction regime best balances task
accuracy and graph recovery, and can offline intervention traces replace online
experimental choice? (RQ3) How do model family, scale, and thinking traces
affect prediction and mechanism recovery across graph sizes? (RQ4) Why do
agents fail, and what simple check can reduce these failures?
The paired prediction and SCM-recovery targets separate task success from
mechanism faithfulness, and DSL traces expose the
hypotheses agents commit to.

\subsection{Experimental Setup}
\label{sec:exp-setup}

\paragraph{Setup.}
The main suite evaluates four models---\texttt{GPT-5-mini},
\texttt{GPT-5.2-high}, \texttt{Qwen3.5-Thinking}, and
\texttt{Qwen3.5-Non-thinking}---on \projectname{}'s 3--7 node graph families,
with up to 50 topologies per (graph size, model) cell and one run
per task. Observation--intervention scaling experiments use
\texttt{GPT-5-mini} and \texttt{GPT-5.2-high} on the 4-node and 6-node suites.
Targeted follow-ups use the 4-/6-node suites, primarily with
\texttt{GPT-5-mini}. All runs use temperature 0.1 and fixed
observation/intervention budgets per graph size
(Appendix~\ref{app:appendix}).
The reactor crystal's hidden
\texttt{frequency}
is the target in every episode, so end-task accuracy is the exact prediction
rate for that value; mechanism recovery is scored
separately with graph, parent, root, edge, and coefficient metrics against the
full episode SCM. Except for the explicit observation--intervention scaling
suite, all follow-up analyses use the mixed regime with two initial
observations; RQ2 motivates this setting as the anchor for subsequent analyses.

\begin{figure}[t]
\centering
\includegraphics[width=0.62\linewidth]{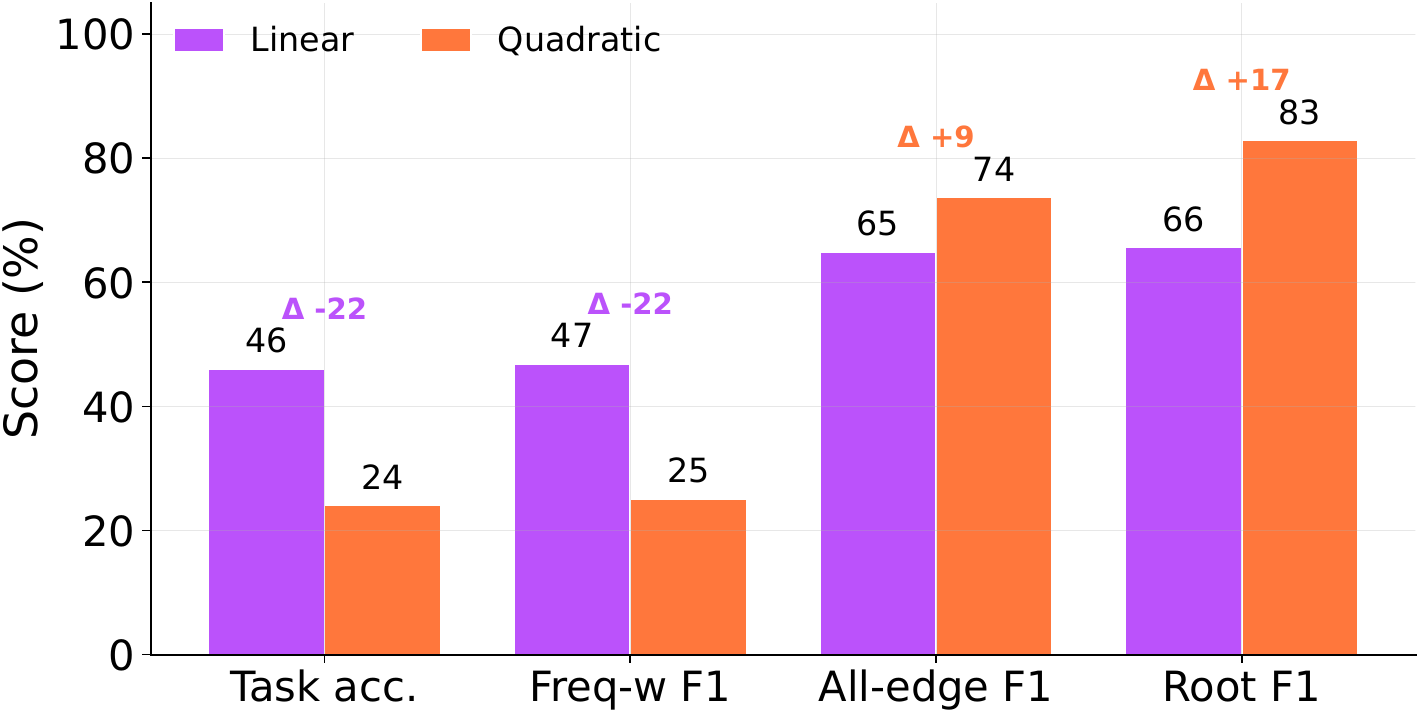}
\caption{Matched 4-node comparison between linear and hard-quadratic
mechanisms for \texttt{GPT-5-mini}. Topology is fixed; only the functional form
changes. Task accuracy and \texttt{frequency}-weight $F_1$ collapse while
all-edge and root-node $F_1$ are preserved or even rise --- agents lose the
mechanism, not the graph.}
\label{fig:linear_vs_quad}
\end{figure}
\subsection{RQ1: Correct Frequency Prediction Does Not Imply Mechanism Recovery}
\label{sec:rq1-dissociation}

\projectname{} pairs each episode with a ground-truth SCM, so we can score the
answer and the mechanism separately. Three controls show that these axes split
in different ways rather than collapsing to one scalar.

\paragraph{Function form.}
Holding the 50 four-node topologies fixed but replacing the linear mechanism
with a hard-quadratic one cuts \texttt{GPT-5-mini} accuracy from 48\% to 24\%
(Figure~\ref{fig:linear_vs_quad}).
The graph is not simply lost: root-node $F_1$ rises (0.559$\to$0.829) and edge
precision is preserved, but \texttt{frequency}-weight $F_1$ collapses
(0.589$\to$0.251; Appendix Table~\ref{tab:linear_quad_metrics}). The agent can
find plausible parents and still fail because it misses the quantitative
mechanism.

\paragraph{Hidden perturbations.}
Off-target hidden noise leaves accuracy near baseline (40--54\% versus 48\%)
but lowers all-edge $F_1$ from 0.79 to 0.61--0.70. When the hidden disturbance
can perturb \texttt{frequency} itself, accuracy drops to 26--40\%
(Appendix Figure~\ref{fig:hidden_variables};
Appendix Table~\ref{tab:hidden_metrics}), showing that some successful
predictions came from fitting a local target equation rather than recovering a
mechanism robust to hidden target perturbations.

\paragraph{Target outgoing edges.}
\texttt{FreqParent} keeps mean edge counts matched but lets
\texttt{frequency} have outgoing edges. Accuracy rises on 4- and 6-node graphs
because the target has fewer incoming edges to fit, while all-edge recovery
falls because global directionality is harder
(Appendix Figure~\ref{fig:FreqParent_followup_metrics};
Appendix Table~\ref{tab:freqparent_followup_metrics}).

\begin{tcolorbox}[takeawaysbox]
Prediction accuracy is necessary but not sufficient evidence of mechanism
recovery.
\end{tcolorbox}

\begin{figure}[t]
\centering
\includegraphics[width=0.58\linewidth]{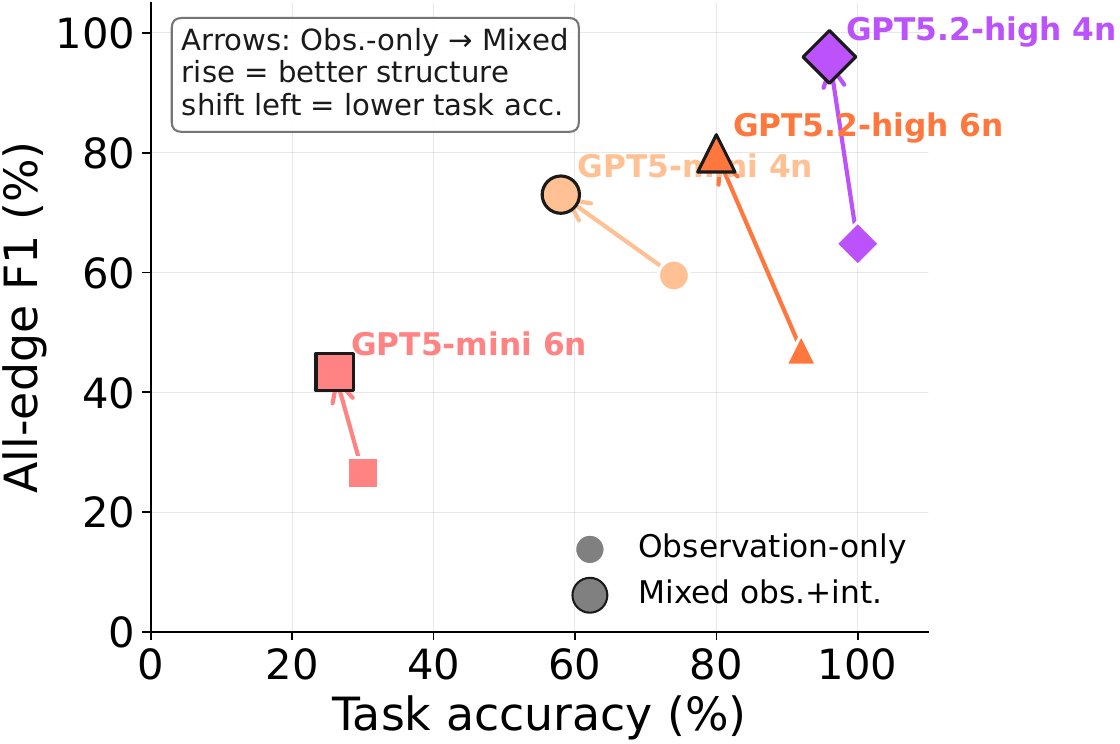}
\caption{Prediction-versus-recovery gap across the four scaling families.
Each suite is shown as an Obs.-only $\to$ Mixed arrow in
(task accuracy, all-edge $F_1$) space: mixed regimes consistently shift
mass toward higher graph fidelity at comparable or better task accuracy.}
\label{fig:disambiguation_gap_summary}
\end{figure}
\subsection{RQ2: Observation-Conditioned Online Intervention Outperforms Pure and Offline Regimes}
\label{sec:rq2-interaction}
RQ2 separates two questions: whether agents need observations, interventions,
or both; and whether offline intervention data is enough when the agent does not choose
the experiments online. Figure~\ref{fig:disambiguation_gap_summary} summarizes
the three online regimes across our four scaling families. For
\texttt{GPT-5-mini}, pure observation often gives the strongest end-task
accuracy on the easier graphs, but mixed observation-conditioned intervention
consistently recovers more faithful graphs on both the 4-node and 6-node
families. In the \texttt{GPT-5.2-high} 6-node setting, for example,
observation-only has higher accuracy than mixed (92\% versus 80\%) but much
lower graph-recovery $F_1$ (0.47 versus 0.80). Pure intervention is weak on
both axes, becoming useful only after observation narrows the hypothesis
space. We therefore use mixed online regimes as the anchor for follow-up controls.
The full regime scatter appears in Appendix Figure~\ref{fig:regime_scatter};
full scaling curves and tables appear in Appendix
Figures~\ref{fig:app_scaling_curves} and~\ref{fig:app_reeval_scaling} and
Appendix~\ref{app:full_experiment_tables}.

\begin{figure}[!ht]
\centering
\begin{minipage}[b]{0.49\linewidth}
\centering
\includegraphics[width=\linewidth]{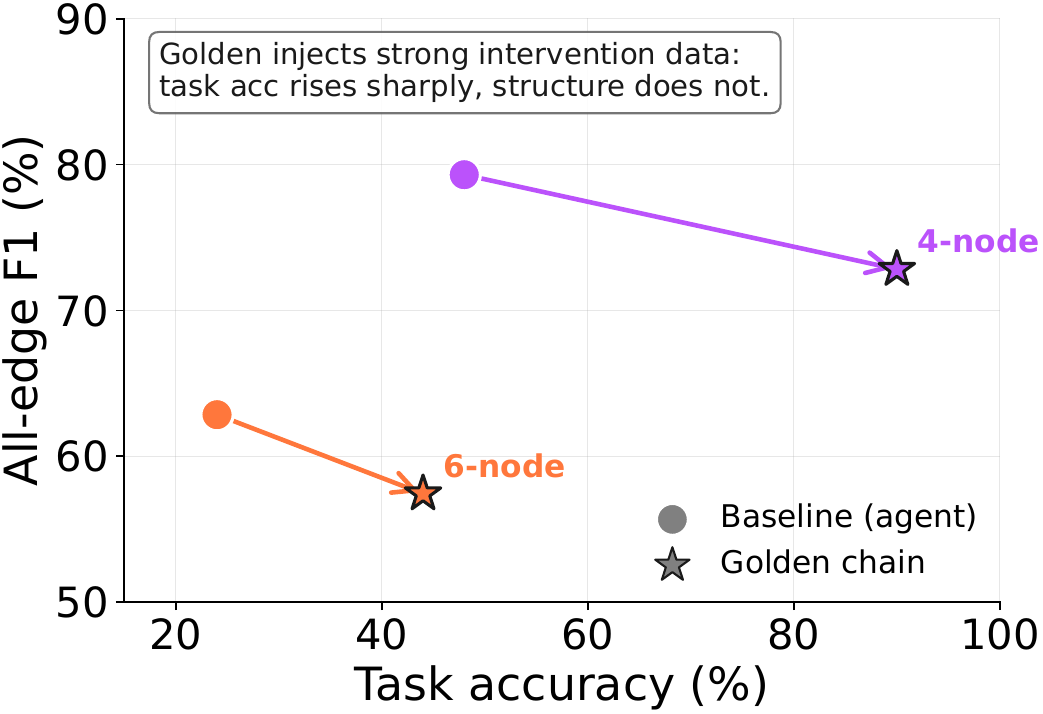}
\captionof{figure}{Golden-intervention experiments on \texttt{GPT-5-mini}.
Baseline $\to$ Golden arrows in (task accuracy, all-edge $F_1$) space:
injected low-MEC intervention traces improve \texttt{frequency} prediction
but hurt all-edge recovery, separating intervention data from intervention
choice.}
\label{fig:golden_followup_metrics}
\end{minipage}\hfill
\begin{minipage}[b]{0.49\linewidth}
\centering
\includegraphics[width=\linewidth]{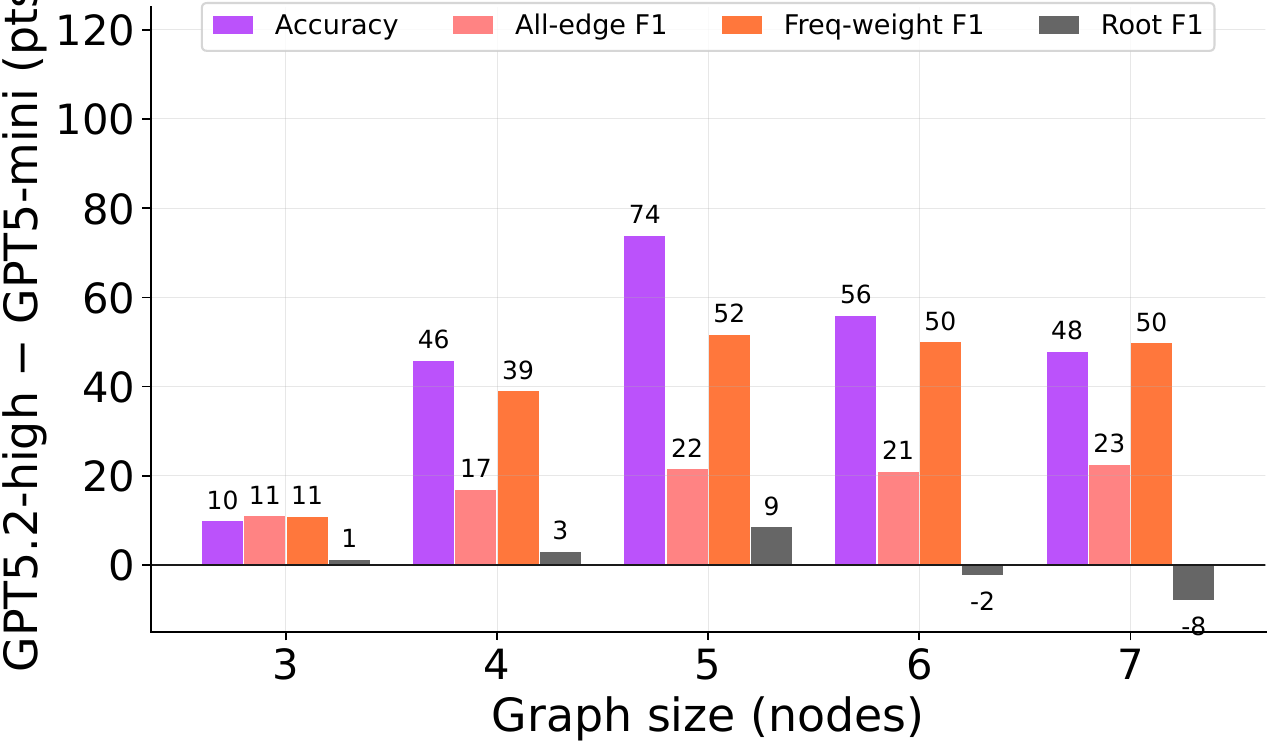}
\captionof{figure}{Capability gap (\texttt{GPT-5.2-high} $-$ \texttt{GPT-5-mini}) in
percentage points across graph sizes and metrics. Scaling concentrates in
accuracy and \texttt{frequency}-weight $F_1$; root-node gains are near zero
or negative at 6--7 nodes, showing where larger models still stall.}
\label{fig:model_metric_support}
\end{minipage}
\end{figure}

The \emph{Golden} control then separates offline intervention data from online
intervention decisions by giving the agent a bounded low-MEC intervention chain
instead of letting it intervene online. Golden improves task accuracy above the
main suite baselines (90\% versus 48\% on 4-node graphs, 44\% versus 24\% on
6-node graphs) but drops all-edge $F_1$ on both sizes
(Figure~\ref{fig:golden_followup_metrics};
Appendix Table~\ref{tab:golden_followup_metrics}). High-quality intervention
chains therefore behave mostly like stronger observations: they help fit the
target equation, but they do not replace the structural signal supplied by the
agent's own online intervention loop.

\begin{tcolorbox}[takeawaysbox]
Observation-conditioned online intervention gives the best balance: observations
narrow the hypothesis space, while agent-chosen interventions recover more
faithful graphs.
\end{tcolorbox}

\subsection{RQ3: Model Family and Scale Pay Off Unevenly Across the Two Axes}
\label{sec:rq3-scaling}

\begin{figure}[!ht]
\centering
\begin{minipage}[t]{0.45\linewidth}
\centering
\includegraphics[width=\linewidth]{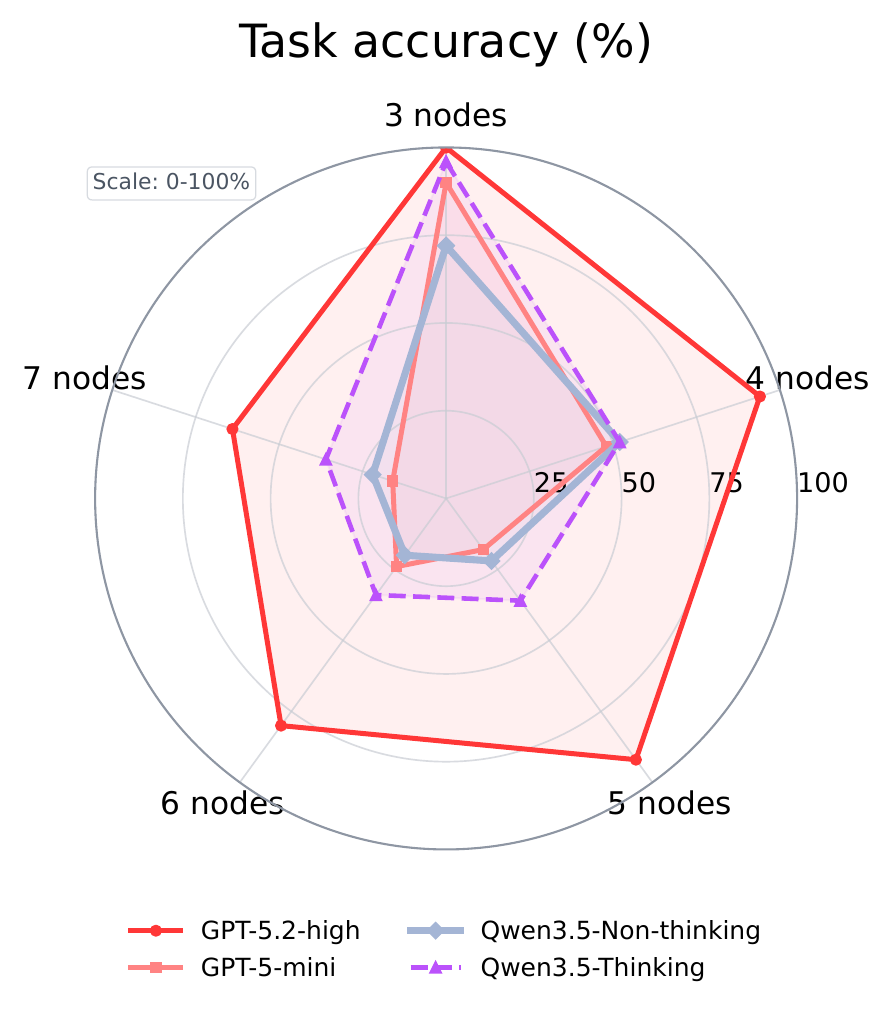}\\
{\footnotesize (a) Task accuracy.}
\end{minipage}\hfill
\begin{minipage}[t]{0.45\linewidth}
\centering
\includegraphics[width=\linewidth]{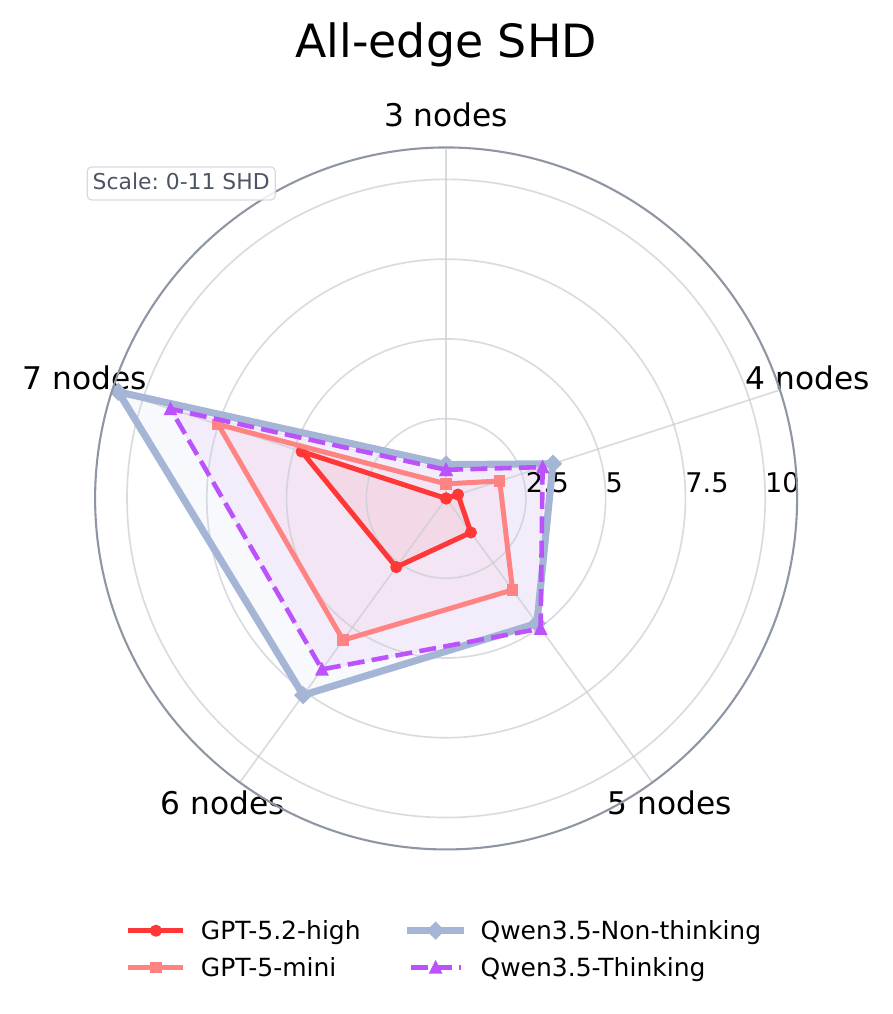}\\
{\footnotesize (b) Directed all-edge SHD.}
\end{minipage}
\caption{Four-model comparison across the 3--7 node main suites. Each vertex
fixes graph size and compares \texttt{GPT-5.2-high},
\texttt{GPT-5-mini}, \texttt{Qwen3.5-Thinking}, and
\texttt{Qwen3.5-Non-thinking}. Lower SHD is better; a reversed edge counts as
one directed SHD error. Task values are endpoint reactor accuracies.}
\label{fig:four_model_node_radar}
\end{figure}

\texttt{GPT-5.2-high} outperforms \texttt{GPT-5-mini} across graph sizes, but
the gains concentrate on mediated structure and quantitative mechanism
fitting rather than every metric uniformly. Figure~\ref{fig:four_model_node_radar}
extends the model-family comparison to all 3--7 node main suites, covering the
two GPT models and \texttt{Qwen3.5} with and without thinking traces.
\texttt{GPT-5.2-high} is the strongest model overall, with the best endpoint
accuracy and lowest directed all-edge SHD at every graph size. Open-weight
\texttt{Qwen3.5} models can be competitive with \texttt{GPT-5-mini} on some
task scores, but their SHD rises faster as graph size grows. Thinking generally
improves Qwen structure recovery, lowering SHD at four graph sizes and raising
all-edge \(F_1\) at every measured size. Across the
full 3--7 node sweep, even \texttt{GPT-5.2-high} still drops to 64\%
accuracy and directed SHD 4.761 at 7 nodes (Figure~\ref{fig:four_model_node_radar}),
and the per-metric gap (Figure~\ref{fig:model_metric_support}) concentrates in
accuracy and \texttt{frequency}-weight $F_1$ while root-node gains flatten on
6--7 node graphs. Absolute metric trajectories appear in Appendix
Figure~\ref{fig:scaling_metric_curves}; per-model metrics are in
Appendix~\ref{app:full_experiment_tables}.

\begin{tcolorbox}[takeawaysbox]
Scaling improves direct-parent and coefficient recovery, but does not remove
the need for better exploration and mechanism-checking methods; thinking helps
Qwen recover structure over most graph sizes, but does not close the gap to
the strongest GPT model.
\end{tcolorbox}

\subsection{RQ4: Agents Fail by Stopping Early, and a Single Verification Step Helps}
\label{sec:rq4-failure}
The DSL trajectories show that failure is often early commitment rather than
missing data. Across the main 4-node and 5-node experiments, both successful
and failed runs leave about half of the intervention budget unused, while more
budget yields only modest gains (Appendix Figure~\ref{fig:overconfidence_diagnostics}).

The DSL lets us
inspect \emph{why} agents stop: at the step where each run commits to a final
answer, we compare the recorded hypothesis $H_t$ (graph + structural
equation + coefficients) against the data already collected. Successful runs
mostly end with hypotheses consistent with their own
$\mathcal{D}_{\le t}$, while failed runs end with hypotheses that mispredict
the very observations and intervention outcomes that produced them.
The
failure mode is therefore better described as
\emph{overconfidence}: agents treat an unverified hypothesis as a final
theory rather than continuing to spend budget on disambiguating experiments.
Consistent with this reading, a single explicit
verification step that checks $H_T$ against $\mathcal{D}_{\le T}$ before
committing raises 4-node accuracy from 48\% to 60\%, making verification a
cheap fix for a trajectory-level failure mode.

\begin{tcolorbox}[takeawaysbox]
Many failures are not caused by an exhausted budget; they are caused by
committing before checking whether the proposed SCM explains the evidence
already collected.
\end{tcolorbox}

\section{Discussion and Conclusion}
\projectname{} should be read as a controlled stress test for interactive causal discovery, not as a broad claim about causal reasoning in arbitrary real-world systems. Its scope is synthetic 3--7 node SCMs, mostly linear mechanisms with one quadratic family, a limited set of model families, and shift-style interventions that adjust variables through the laboratory interface rather than implement perfect hard-\(do\) operations. Within that scope, the benchmark shows that agents can collect useful evidence and predict the held-out \texttt{frequency} while still recovering an incomplete or wrong mechanism. By scoring both final prediction and the evolving SCM hypothesis, \projectname{} exposes where scale helps and where graph complexity, quantitative mechanism fitting, and premature hypothesis commitment remain central bottlenecks.

\section{Potential Risks}

\projectname{} is a synthetic benchmark, so its results should not be read as
evidence that an agent is ready for real scientific, medical, or policy
deployment. The main risk is overgeneralizing success on templated SCM tasks to
high-stakes settings with real interventions and domain constraints.

\section{Limitations}

\projectname{} is a controlled synthetic benchmark, so its results should be
interpreted within that scope. The current suites use 3--7 node SCMs with
mostly linear mechanisms and one hard-quadratic extension, so they do not cover
the full range of causal structures, functional forms, latent variables, or
measurement processes found in real scientific domains.

Our experiments cover a finite set of LLM agents, prompts, and interaction
budgets. Performance may differ with other models, tool interfaces, decoding
policies, or longer exploration budgets, and our main analyses focus on final
predictions and final hypotheses rather than every possible trajectory-level
diagnostic.

\section*{Acknowledgements}
This work is in part supported by the University of Chicago Novel Intelligence Research Initiative and AI research pillars, 
NSF Grants IIS-2126602, IIS-2302785, CHE-2505932, an Amazon AICE Award, gift funding from AI2, and a grant from Coefficient Giving.

\small
\bibliography{references}

\begin{thebibliography}{39}
\providecommand{\natexlab}[1]{#1}
\providecommand{\url}[1]{\texttt{#1}}
\expandafter\ifx\csname urlstyle\endcsname\relax
  \providecommand{\doi}[1]{doi: #1}\else
  \providecommand{\doi}{doi: \begingroup \urlstyle{rm}\Url}\fi

\bibitem[Abdulaal et~al.(2024)Abdulaal, adamos hadjivasiliou, Montana-Brown, He, Ijishakin, Drobnjak, Castro, and Alexander]{abdulaal2024causalmodellingagents}
Ahmed Abdulaal, adamos hadjivasiliou, Nina Montana-Brown, Tiantian He, Ayodeji Ijishakin, Ivana Drobnjak, Daniel~C. Castro, and Daniel~C. Alexander.
\newblock Causal modelling agents: Causal graph discovery through synergising metadata- and data-driven reasoning.
\newblock In \emph{The Twelfth International Conference on Learning Representations}, 2024.
\newblock URL \url{https://openreview.net/forum?id=pAoqRlTBtY}.

\bibitem[Andersson et~al.(1997)Andersson, Madigan, and Perlman]{andersson1997mec}
Steen~A. Andersson, David Madigan, and Michael~D. Perlman.
\newblock {A characterization of Markov equivalence classes for acyclic digraphs}.
\newblock \emph{The Annals of Statistics}, 25\penalty0 (2):\penalty0 505 -- 541, 1997.
\newblock \doi{10.1214/aos/1031833662}.
\newblock URL \url{https://doi.org/10.1214/aos/1031833662}.

\bibitem[Chen et~al.(2024{\natexlab{a}})Chen, Peng, Chen, Wang, Xu, Zeng, Zhao, Zhao, Qiao, and Lu]{chen2024causal}
Sirui Chen, Bo~Peng, Meiqi Chen, Ruiqi Wang, Mengying Xu, Xingyu Zeng, Rui Zhao, Shengjie Zhao, Yu~Qiao, and Chaochao Lu.
\newblock Causal evaluation of language models, 2024{\natexlab{a}}.

\bibitem[Chen et~al.(2024{\natexlab{b}})Chen, Xu, Wang, Zeng, Zhao, Zhao, and Lu]{clear2024}
Sirui Chen, Mengying Xu, Kun Wang, Xingyu Zeng, Rui Zhao, Shengjie Zhao, and Chaochao Lu.
\newblock Clear: Can language models really understand causal graphs?, 2024{\natexlab{b}}.
\newblock URL \url{https://arxiv.org/abs/2406.16605}.

\bibitem[Chen et~al.(2025)Chen, Anumasa, Lin, Shah, Goyal, and Liu]{chen2025autobench}
Tingting Chen, Srinivas Anumasa, Beibei Lin, Vedant Shah, Anirudh Goyal, and Dianbo Liu.
\newblock {Auto-Bench}: An automated benchmark for scientific discovery in {LLMs}, 2025.
\newblock URL \url{https://arxiv.org/abs/2502.15224}.

\bibitem[Chen et~al.(2026)Chen, Chen, Liu, Yu, Song, Li, Li, Torr, Han, and Zhang]{chen2026causalgame}
Zhenhao Chen, Yongqiang Chen, Chenxi Liu, Junchi Yu, Xiangchen Song, Zijian Li, Jialin Li, Philip Torr, Bo~Han, and Kun Zhang.
\newblock Causalgame: Benchmarking causal thinking of llm agents in games.
\newblock In \emph{ICLR 2026 Workshop on Foundation Models for Science}, 2026.
\newblock URL \url{https://openreview.net/forum?id=SEFSkn4l6d}.

\bibitem[Darvariu et~al.(2024)Darvariu, Hailes, and Musolesi]{llmeffectivepriors2024}
Victor{-}Alexandru Darvariu, Stephen Hailes, and Mirco Musolesi.
\newblock Large language models are effective priors for causal graph discovery, 2024.
\newblock URL \url{https://arxiv.org/abs/2405.13551}.

\bibitem[Dunbar and Fugelsang(2005)]{dunbar_fugelsang_2005}
K.~Dunbar and J.~Fugelsang.
\newblock Causal thinking in science: How scientists and students interpret the unexpected.
\newblock In M.~E. Gorman, R.~D. Tweney, D.~C. Gooding, and A.~P. Kincannon, editors, \emph{Scientific and Technological Thinking}, pages 57--79. Lawrence Erlbaum Associates, Mahwah, NJ, 2005.

\bibitem[Geng et~al.(2025)Geng, Chen, Arumugam, and Griffiths]{geng2025reliableaiscientists}
Jiayi Geng, Howard Chen, Dilip Arumugam, and Thomas~L. Griffiths.
\newblock Are large language models reliable {AI} scientists? assessing reverse-engineering of black-box systems, 2025.
\newblock URL \url{https://arxiv.org/abs/2505.17968}.

\bibitem[Gregorini et~al.(2025)Gregorini, Boldrini, and Valerio]{gregorini2025dodo}
Matteo Gregorini, Chiara Boldrini, and Lorenzo Valerio.
\newblock {DODO}: Causal structure learning with budgeted interventions, 2025.
\newblock URL \url{https://arxiv.org/abs/2510.08207}.

\bibitem[Hauser and B{\"u}hlmann(2012)]{JMLR:v13:hauser12a}
Alain Hauser and Peter B{\"u}hlmann.
\newblock Characterization and greedy learning of interventional markov equivalence classes of directed acyclic graphs.
\newblock \emph{Journal of Machine Learning Research}, 13\penalty0 (79):\penalty0 2409--2464, 2012.
\newblock URL \url{http://jmlr.org/papers/v13/hauser12a.html}.

\bibitem[Havrilla et~al.(2025)Havrilla, Alvarez{-}Melis, and Fusi]{igda2025}
Alex Havrilla, David Alvarez{-}Melis, and Nicolo Fusi.
\newblock Igda: Interactive graph discovery through large language model agents, 2025.
\newblock URL \url{https://arxiv.org/abs/2502.17189}.

\bibitem[Imbens and Rubin(2015)]{imbens_rubin_causal_2015}
Guido~W. Imbens and Donald~B. Rubin.
\newblock \emph{Causal Inference for Statistics, Social, and Biomedical Sciences}.
\newblock Cambridge University Press, 2015.
\newblock ISBN 978-0521885884.
\newblock \doi{10.1017/CBO9781139025751}.
\newblock URL \url{https://doi.org/10.1017/CBO9781139025751}.

\bibitem[Jansen et~al.(2024)Jansen, C{\^{o}}t{\'{e}}, Khot, Bransom, Mishra, Majumder, Tafjord, and Clark]{DBLP:journals/corr/abs-2406-06769}
Peter~A. Jansen, Marc{-}Alexandre C{\^{o}}t{\'{e}}, Tushar Khot, Erin Bransom, Bhavana~Dalvi Mishra, Bodhisattwa~Prasad Majumder, Oyvind Tafjord, and Peter Clark.
\newblock {DISCOVERYWORLD:} {A} virtual environment for developing and evaluating automated scientific discovery agents.
\newblock \emph{CoRR}, abs/2406.06769, 2024.
\newblock \doi{10.48550/ARXIV.2406.06769}.
\newblock URL \url{https://doi.org/10.48550/arXiv.2406.06769}.

\bibitem[Jiang et~al.(2024)Jiang, Sorensen, Levine, and Choi]{jiang2024can}
Liwei Jiang, Taylor Sorensen, Sydney Levine, and Yejin Choi.
\newblock Can language models reason about individualistic human values and preferences?
\newblock \emph{arXiv:2410.03868}, 2024.

\bibitem[Jin et~al.(2023{\natexlab{a}})Jin, Chen, Leeb, Gresele, Kamal, LYU, Blin, Gonzalez~Adauto, Kleiman-Weiner, Sachan, and Sch\"{o}lkopf]{jin2024cladderassessingcausalreasoning}
Zhijing Jin, Yuen Chen, Felix Leeb, Luigi Gresele, Ojasv Kamal, Zhiheng LYU, Kevin Blin, Fernando Gonzalez~Adauto, Max Kleiman-Weiner, Mrinmaya Sachan, and Bernhard Sch\"{o}lkopf.
\newblock Cladder: Assessing causal reasoning in language models.
\newblock In A.~Oh, T.~Naumann, A.~Globerson, K.~Saenko, M.~Hardt, and S.~Levine, editors, \emph{Proceedings of the Advances in Neural Information Processing Systems}, 2023{\natexlab{a}}.
\newblock \doi{10.48550/ARXIV.2312.04350}.
\newblock URL \url{https://doi.org/10.48550/arXiv.2312.04350}.

\bibitem[Jin et~al.(2023{\natexlab{b}})Jin, Liu, Lyu, Poff, Sachan, Mihalcea, Diab, and Sch{\"o}lkopf]{corr2cause2024}
Zhijing Jin, Jiarui Liu, Zhiheng Lyu, Spencer Poff, Mrinmaya Sachan, Rada Mihalcea, Mona Diab, and Bernhard Sch{\"o}lkopf.
\newblock Can large language models infer causation from correlation?, 2023{\natexlab{b}}.
\newblock URL \url{https://arxiv.org/abs/2306.05836}.

\bibitem[Jiralerspong et~al.(2024)Jiralerspong, Chen, More, Shah, and Bengio]{efficientcausalgraph2024}
Thomas Jiralerspong, Xiaoyin Chen, Yash More, Vedant Shah, and Yoshua Bengio.
\newblock Efficient causal graph discovery using large language models, 2024.
\newblock URL \url{https://arxiv.org/abs/2402.01207}.

\bibitem[K{\i}c{\i}man et~al.(2023)K{\i}c{\i}man, Ness, Sharma, and Tan]{kiciman2023causal}
Emre K{\i}c{\i}man, Robert Ness, Amit Sharma, and Chenhao Tan.
\newblock Causal reasoning and large language models: Opening a new frontier for causality, 2023.
\newblock URL \url{https://arxiv.org/abs/2305.00050}.

\bibitem[Lampinen et~al.(2023)Lampinen, Chan, Dasgupta, Nam, and Wang]{lampinen2023passive}
Andrew~K. Lampinen, Stephanie C.~Y. Chan, Ishita Dasgupta, Andrew~J. Nam, and Jane~X. Wang.
\newblock Passive learning of active causal strategies in agents and language models.
\newblock In \emph{Proceedings of the Advances in Neural Information Processing Systems}, 2023.

\bibitem[Langley(2019)]{langley2019scientific}
Pat Langley.
\newblock Scientific discovery, causal explanation, and process model induction.
\newblock \emph{Mind \& Society}, 18\penalty0 (1):\penalty0 43--56, 2019.
\newblock \doi{10.1007/s11299-019-00216-1}.
\newblock URL \url{https://doi.org/10.1007/s11299-019-00216-1}.

\bibitem[Liu et~al.(2025)Liu, Huang, Hu, Zhou, and Tan]{liu2025hypobench}
Haokun Liu, Sicong Huang, Jingyu Hu, Yangqiaoyu Zhou, and Chenhao Tan.
\newblock Hypobench: Towards systematic and principled benchmarking for hypothesis generation.
\newblock \emph{arXiv:2504.11524}, 2025.

\bibitem[Liu et~al.(2023)Liu, Zhang, Guo, Jin, Li, Wei, and Sun]{liu2023kept}
Jintao Liu, Zequn Zhang, Zhi Guo, Li~Jin, Xiaoyu Li, Kaiwen Wei, and Xian Sun.
\newblock Kept: Knowledge enhanced prompt tuning for event causality identification.
\newblock \emph{Knowledge\-Based Systems}, 259, 2023.

\bibitem[Long et~al.(2023)Long, Schuster, and Pich{\'{e}}]{llmcausalgraphs2023}
Stephanie Long, Tibor Schuster, and Alexandre Pich{\'{e}}.
\newblock Can large language models build causal graphs?, 2023.
\newblock URL \url{https://arxiv.org/abs/2303.05279}.

\bibitem[Mooij et~al.(2020)Mooij, Magliacane, and Claassen]{JMLR:v21:17-123}
Joris~M. Mooij, Sara Magliacane, and Tom Claassen.
\newblock Joint causal inference from multiple contexts.
\newblock \emph{Journal of Machine Learning Research}, 21\penalty0 (99):\penalty0 1--108, 2020.
\newblock URL \url{http://jmlr.org/papers/v21/17-123.html}.

\bibitem[Pearl(2009)]{Pearl_2009}
Judea Pearl.
\newblock \emph{Causality: Models, Reasoning, and Inference}.
\newblock Cambridge University Press, 2 edition, sep 2009.
\newblock ISBN 9780511803161.
\newblock \doi{10.1017/cbo9780511803161}.
\newblock URL \url{https://doi.org/10.1017/CBO9780511803161}.

\bibitem[Pearl and Mackenzie(2018)]{pearl2018why}
Judea Pearl and Dana Mackenzie.
\newblock \emph{The Book of Why: The New Science of Cause and Effect}.
\newblock Basic Books, Inc., USA, 1st edition, 2018.
\newblock ISBN 046509760X.

\bibitem[Qin et~al.(2019)Qin, Bosselut, Holtzman, Bhagavatula, Clark, and Choi]{qin2019counterfactual}
Lianhui Qin, Antoine Bosselut, Ari Holtzman, Chandra Bhagavatula, Elizabeth Clark, and Yejin Choi.
\newblock Counterfactual story reasoning and generation.
\newblock In \emph{Proceedings of the 2019 Conference on Empirical Methods in Natural Language Processing and the 9th International Joint Conference on Natural Language Processing (EMNLP-IJCNLP)}, pages 5043--5053, Hong Kong, China, November 2019. Association for Computational Linguistics.
\newblock \doi{10.18653/v1/D19-1509}.
\newblock URL \url{https://aclanthology.org/D19-1509/}.

\bibitem[Romanou et~al.(2023)Romanou, Montariol, Paul, Laugier, Aberer, and Bosselut]{romanou2023crab}
Angelika Romanou, Syrielle Montariol, Debjit Paul, Léo Laugier, Karl Aberer, and Antoine Bosselut.
\newblock Crab: Assessing the strength of causal relationships between real‑world events.
\newblock In \emph{Proceedings of the 2023 Conference on Empirical Methods in Natural Language Processing (EMNLP 2023)}, pages 15198--15216, 2023.
\newblock \doi{10.18653/V1/2023.EMNLP-MAIN.940}.
\newblock URL \url{https://aclanthology.org/2023.emnlp-main.940.pdf}.

\bibitem[Rothenh{\"a}usler et~al.(2015)Rothenh{\"a}usler, Heinze, Peters, and Meinshausen]{backshift2015}
Dominik Rothenh{\"a}usler, Christina Heinze, Jonas Peters, and Nicolai Meinshausen.
\newblock backshift: Learning causal cyclic graphs from unknown shift interventions, 2015.
\newblock URL \url{https://arxiv.org/abs/1506.02494}.

\bibitem[Stolfo et~al.(2023)Stolfo, Jin, Shridhar, Schölkopf, and Sachan]{stolfo2023causal}
Alessandro Stolfo, Zhijing Jin, Kumar Shridhar, Bernhard Schölkopf, and Mrinmaya Sachan.
\newblock A causal framework to quantify the robustness of mathematical reasoning with language models.
\newblock In \emph{Proceedings of the 61st Annual Meeting of the Association for Computational Linguistics}. Association for Computational Linguistics, 2023.
\newblock URL \url{https://arxiv.org/abs/2210.12023}.

\bibitem[Vashishtha et~al.(2023)Vashishtha, Reddy, Kumar, Bachu, Balasubramanian, and Sharma]{causalorder2023}
Aniket Vashishtha, Abbavaram~Gowtham Reddy, Abhinav Kumar, Saketh Bachu, Vineeth~N Balasubramanian, and Amit Sharma.
\newblock Causal order: The key to leveraging imperfect experts in causal inference, 2023.
\newblock URL \url{https://arxiv.org/abs/2310.15117}.

\bibitem[Vashishtha et~al.(2025)Vashishtha, Kumar, Pandey, Reddy, Ahuja, Balasubramanian, and Sharma]{vashishtha2025teachingtransformerscausalreasoning}
Aniket Vashishtha, Abhinav Kumar, Atharva Pandey, Abbavaram~Gowtham Reddy, Kabir Ahuja, Vineeth~N Balasubramanian, and Amit Sharma.
\newblock Teaching transformers causal reasoning through axiomatic training.
\newblock In \emph{Proceedings of the International Conference on Machine Learning}, 2025.

\bibitem[Wang(2024)]{wang-2024-causalbench}
Zeyu Wang.
\newblock {C}ausal{B}ench: A comprehensive benchmark for evaluating causal reasoning capabilities of large language models.
\newblock In \emph{Proceedings of the 10th SIGHAN Workshop on Chinese Language Processing (SIGHAN-10)}, pages 143--151, Bangkok, Thailand, August 2024. Association for Computational Linguistics.
\newblock URL \url{https://aclanthology.org/2024.sighan-1.17/}.

\bibitem[Yamaoka et~al.(2026)Yamaoka, Mukherjee, G{\"a}rtner, Selby, Konigorski, H{\"u}llermeier, Bengs, and Vollmer]{yamaoka2026linearllmscm}
Kanta Yamaoka, Sumantrak Mukherjee, Thomas G{\"a}rtner, David~Antony Selby, Stefan Konigorski, Eyke H{\"u}llermeier, Viktor Bengs, and Sebastian~Josef Vollmer.
\newblock {Linear-LLM-SCM}: Benchmarking {LLMs} for coefficient elicitation in linear-gaussian causal models, 2026.
\newblock URL \url{https://arxiv.org/abs/2602.10282}.

\bibitem[Yao et~al.(2023)Yao, Zhao, Yu, Du, Shafran, Narasimhan, and Cao]{DBLP:conf/iclr/YaoZYDSN023}
Shunyu Yao, Jeffrey Zhao, Dian Yu, Nan Du, Izhak Shafran, Karthik~R. Narasimhan, and Yuan Cao.
\newblock React: Synergizing reasoning and acting in language models.
\newblock In \emph{The Eleventh International Conference on Learning Representations, {ICLR} 2023, Kigali, Rwanda, May 1-5, 2023}. OpenReview.net, 2023.
\newblock URL \url{https://openreview.net/forum?id=WE_vluYUL-X}.

\bibitem[Zang et~al.(2023)Zang, Wang, Pei, and Liang]{zang2023discovering}
Chuanqi Zang, Hanqing Wang, Mingtao Pei, and Wei Liang.
\newblock Discovering the real association: Multimodal causal reasoning in video question answering.
\newblock In \emph{Proceedings of the IEEE/CVF Conference on Computer Vision and Pattern Recognition (CVPR)}, pages 19027--19036, Vancouver, Canada, June 2023.
\newblock \doi{10.1109/CVPR52729.2023.01824}.
\newblock URL \url{https://openaccess.thecvf.com/content/CVPR2023/html/Zang_Discovering_the_Real_Association_Multimodal_Causal_Reasoning_in_Video_Question_CVPR_2023_paper.html}.

\bibitem[Ze{\v{c}}evi{\'c} et~al.(2023)Ze{\v{c}}evi{\'c}, Willig, Dhami, and Kersting]{zecevic2023causal}
Matej Ze{\v{c}}evi{\'c}, Moritz Willig, Devendra~Singh Dhami, and Kristian Kersting.
\newblock Causal parrots: Large language models may talk causality but are not causal.
\newblock \emph{Transactions in Machine Learning Research}, 2023.
\newblock \doi{10.48550/ARXIV.2308.13067}.
\newblock URL \url{https://arxiv.org/abs/2308.13067}.

\bibitem[Zheng et~al.(2023)Zheng, Ma, Qiu, Wu, Ma, Liu, Feng, Shang, and Chen]{zheng2023preservingcommonsenseknowledgepretrained}
Junhao Zheng, Qianli Ma, Shengjie Qiu, Yue Wu, Peitian Ma, Junlong Liu, Huawen Feng, Xichen Shang, and Haibin Chen.
\newblock Preserving commonsense knowledge from pre-trained language models via causal inference.
\newblock In Anna Rogers, Jordan Boyd-Graber, and Naoaki Okazaki, editors, \emph{Proceedings of the 61st Annual Meeting of the Association for Computational Linguistics (Volume 1: Long Papers)}, pages 9155--9173, Toronto, Canada, July 2023. Association for Computational Linguistics.
\newblock \doi{10.18653/v1/2023.acl-long.509}.
\newblock URL \url{https://aclanthology.org/2023.acl-long.509/}.

\end{thebibliography}

\newpage

\appendix

\etocsetlocaltop.toc{part}
\etocsetnexttocdepth{subsection}
\localtableofcontents
\clearpage

\section{Appendix}
\label{app:appendix}

\subsection{Benchmark Setup and Causal-Discovery Context}
\label{app:setup_details}

This section collects setup details that are needed to interpret the main
results but are too mechanical for the main text. The benchmark uses a fixed
budget policy except in the explicit observation--intervention scaling suites:
for a \(k\)-node graph, the observation budget is 2 and the intervention budget
is \(4(k - 1)\). The graph families span 3--7 node SCMs; Table~\ref{tab:app_main_graph_stats}
summarizes the topology distribution used in the main experiments.

Classical causal discovery studies how causal structure can be learned from
observational, interventional, or shifted data, including constraint- and
score-based discovery, interventional Markov equivalence, discovery across
multiple contexts, and unknown shift interventions
\citep{Pearl_2009,andersson1997mec,JMLR:v13:hauser12a,JMLR:v21:17-123,backshift2015,zang2023discovering}.
\projectname{} is closest to the shift-intervention regime of this literature,
but it does not propose a new discovery algorithm or assume perfect
\(do\)-interventions. Instead, it evaluates whether an LLM agent can recover
the graph and equations of a hidden SCM through a finite sequence of
shift-style interventions on controllable properties.

\subsection{SCM and Hidden-Disturbance Details}
\label{app:scm_details}

Formally, each episode instantiates an SCM
\(\mathcal{M}=(\mathbf{U},\mathbf{V},F,P(\mathbf{U}))\)
\citep{Pearl_2009}: \(\mathbf{V}\) are endogenous variables determined inside
the system, \(\mathbf{U}\) are exogenous variables drawn from \(P(\mathbf{U})\),
and \(F\) is a collection of structural equations mapping each variable's
parents and exogenous term to its value. In \projectname{},
\(\mathbf{V}=O\cup\{Y\}\), where \(O\) are observable property variables and
\(Y=\texttt{frequency}\). Root variables remain endogenous nodes, but their
values are generated from exogenous source terms; in hidden-noise suites, the
exogenous terms also include an unobserved disturbance \(H\).

The agent observes \(O\) and \(Y\) in the prior evidence records, may observe
and intervene on the configured subset \(C\subseteq O\) of the manipulator
crystal during interaction, and observes only \(O\) on the reactor crystal.
Variables in \(O\setminus C\) are observable but not controllable; \(Y\) is
never controllable; and \(H\), when present, is neither observable nor
controllable. Hidden-disturbance suites resample \(H\) after each intervention
and add it as a fixed-weight shift to a designated subset of observable
endogenous variables; these shifted values then propagate downstream through
the structural equations. The agent sees only the resulting observed values,
not \(H\) itself.

\begin{table}[t]
\centering
\small
\setlength{\tabcolsep}{4pt}
\begin{tabular}{lccccc}
\toprule
Nodes & Graphs & Edge mean & Edge var. & Fork mean & Collider mean \\
\midrule
3 & 50 & 2.56 & 0.25 & 0.68 & 0.72 \\
4 & 50 & 4.54 & 1.01 & 2.18 & 2.22 \\
5 & 50 & 7.26 & 4.23 & 5.54 & 5.26 \\
6 & 50 & 8.82 & 5.15 & 6.72 & 6.22 \\
7 & 50 & 10.26 & 4.95 & 7.84 & 7.00 \\
\bottomrule
\end{tabular}
\caption{Main-experiment graph statistics. Forks and colliders are counted as three-node motif instances; edge variance is the population variance over the 50 graphs in each row.}
\label{tab:app_main_graph_stats}
\end{table}

\begin{table}[t]
\centering
\small
\setlength{\tabcolsep}{4pt}
\begin{tabular}{lccc}
\toprule
\textbf{Variable role} & \textbf{SCM role} & \textbf{Observable} & \textbf{Intervenable} \\
\midrule
Controllable property ($X\in C$) & Endogenous & \checkmark & \checkmark \\
Non-controllable property ($X\in O\setminus C$) & Endogenous & \checkmark & --- \\
Target \(Y=\texttt{frequency}\) (evidence records / manipulator crystal) & Endogenous & \checkmark & --- \\
Target \(Y=\texttt{frequency}\) (reactor crystal) & Endogenous & --- & --- \\
Hidden disturbance \(H\) (when present) & Exogenous & --- & --- \\
\bottomrule
\end{tabular}
\caption{Per-variable SCM role and access for the agent in a \projectname{} episode. The agent observes prior evidence records, intervenes only on the configured controllable subset \(C\) of the manipulator crystal, never observes hidden exogenous disturbances, and predicts the held-out \texttt{frequency} of the reactor crystal.}
\label{tab:variable_access}
\end{table}

\subsection{Artifact and Implementation Details}
\label{app:artifact_implementation}

\projectname{} builds its embodied interface on DiscoveryWorld
\citep{DBLP:journals/corr/abs-2406-06769}, released under the Apache-2.0
license; its SCM generator, intervention/reactor mechanics, DSL traces, and
scoring code are new synthetic research and evaluation artifacts. This use is
consistent with DiscoveryWorld's role as a virtual environment for evaluating
scientific-discovery agents. The benchmark is generated from fixed templates
over synthetic laboratory variables, so it contains no demographic attributes,
personal identifiers, or naturally occurring offensive web text. Appendix
\ref{app:scm_details} and \ref{app:full_experiment_tables} document artifact
coverage, variable access, and evaluation/experiment settings.

\subsection{Prompt Templates}

The benchmark uses a two-stage prompting scheme. The first prompt governs
iterative hypothesis--experiment interaction in the environment, while the
second prompt frames the reactor activation objective and embeds the
task-specific laboratory rules. The templates below are the exact prompt files
used in the implementation, with runtime placeholders filled by the environment
during evaluation.

\begin{tcblisting}{
  promptbox,
  title={Phase 1 Controller Prompt},
  listing only,
  listing options={
    basicstyle=\ttfamily\scriptsize,
    breaklines=true,
    columns=fullflexible,
    keepspaces=true,
    showstringspaces=false
  }
}
You are playing a video game about making scientific discoveries.  The game is in the style of a 2D top-down RPG (you are the agent in the center of the image), and as input you get both an image, as well as information from the user interface (provided in the JSON below) that describes your location, inventory, objects in front of you, the result of your last action, and the task that you're assigned to complete.
Because this is a game, the actions that you can complete are limited to a set of actions that are defined by the game. Those are also described below.
This game is played step-by-step.  At each step, you get the input that I am providing, and output a single action to take as the next step.

Navigation note: 
Moving forward moves you in the direction you're facing. You are currently facing "{{ facing_direction }}"`. From your current location, the directions that you can move to (i.e. they don't have an object blocking them) are: {{ valid_dirs }}. 

Interaction note:
You can only interact (i.e. take actions with) objects that are in your inventory, or directly (i.e. one square) in front of you, in the direction that you're facing.  E.g. if you want to pick an object up, you need to move directly in front of it, and face it, before using the pick-up action on it.

Teleportation note: 
To make moving easier, you can teleport to a list of specific locations in the environment, using the teleport action.  In this case, 'arg1' is the name of a location. An example teleport action would be: `{"action": "TELEPORT_TO_LOCATION", "arg1": "school"}.


Action Format note:
{{ additional_instructions }}

**Required Action Structure** (no memory or thought fields):

Your response MUST be a single JSON object with keys in this exact order: memory -> thought -> past_data -> hypothesis -> experiment -> (action fields)

** CRITICAL: INCREMENTAL UPDATE RULE **
You will see a "Previous State" section showing your last `past_data` and `hypothesis` from the previous step.
- **ALWAYS start from the previous state and make INCREMENTAL updates**
- **DO NOT discard previous data or reset known coefficients to null**
- For `past_data`: COPY all previous entries and APPEND new observations (do not recreate from scratch)
- For `hypothesis`: COPY all previous edges/coefficients and UPDATE only what changed based on new evidence
- Only update when there is NEW evidence; otherwise keep the previous values unchanged

1. **memory** :a concise but comprehensive running summary that you update EVERY step using (a) your last memory and (b) the most recent few action-observation pairs.
   - Keep memory as complete as possible without being verbose or repeating low-level noise.
   - Include: current subgoal and plan, useful facts discovered, items/locations that matter, pending checks/unknowns, recent failures and why...
   - Do not list raw logs; summarize to help immediately guide the next step.
   - When you observe measurement data, property values, or frequency readings, you should remember them in your memory section"
   - This helps you track patterns, relationships, and make informed decisions"
   - Record key observations such as: crystal properties, frequency measurements, causal relationships you discover, and experimental results"
2. **thought** (string): Natural language explanation of why the next action is chosen given current data and hypothesis.
   - Should reference specific gaps in data or hypothesis that motivate the action
   - MUST include accessibility checks: explicitly verify that any UUIDs you reference appear in the interactable objects list or your inventory, and state which list (inventory or interactable)
   - If previous actions failed, explain why they failed and justify why the new action will succeed
   - If transitioning between phases (e.g., property manipulator to reactor), explain the transition logic

3. **past_data** (JSON array): Records all available evidence, including baseline, passive observations, and intervention results.
   - Structure: `[{"id":"0/N","props":{"pH":95,"Pressure":103,...},"freq":610}, {"id":"1/N","props":{"pH":95,"Pressure":50,...},"freq":451}, ...]`
   - Each entry MUST include: `id` (experiment counter), `props` (all measured properties as key-value pairs), `freq` (resonance frequency)
   - If available, the first entry (id="0/N") is the baseline before any interventions
   - **INCREMENTAL UPDATE**: Look at "Last past_data" in Previous State -> COPY ALL existing entries -> APPEND new observations
   - Update rule: APPEND whenever you obtain a new informative observation, whether from passive evidence or from a property-manipulator intervention; keep unchanged during reactor operations unless genuinely new evidence appears there
   - **DO NOT recreate the array from scratch; always preserve all previous entries**

4. **hypothesis** (JSON object): Current understanding of causal structure and frequency relationships.
   - Structure: `{"edges":[{"from":"PropA","to":"PropB"},...], "freq_equation":"resonanceFreq = base + c_PropA*PropA + c_PropB*PropB", "coefficients":{"base":value,"c_PropA":value,"c_PropB":value}}`
   - `edges`: Directed causal relationships between properties(including frequency)
   - `freq_equation`: String representation of hypothesized frequency formula
   - `coefficients`: Numerical values for the equation (use null for unknown coefficients, use real value for known coefficients)
   - **CRITICAL NAMING CONVENTIONS**:
     * Coefficient names MUST follow the pattern `c_{property_name}` (e.g., `c_temperatureC`, `c_conductivity`)
     * ONLY include properties in freq_equation and coefficients that have edges pointing TO the frequency property
     * DO NOT include coefficients with value 0 (except "base") - if a property doesn't affect frequency, exclude it entirely
     * Example: If edges show `temperatureC->resonanceFreq` and `conductivity->resonanceFreq`, and coefficients calculated as c_temperatureC=2 and c_conductivity=1, then:
       - freq_equation: `"resonanceFreq = base + c_temperatureC*temperatureC + c_conductivity*conductivity"`
       - coefficients: `{"base":20, "c_temperatureC":2, "c_conductivity":1}` (no other properties)
   - **INCREMENTAL UPDATE**: Look at "Last hypothesis" in Previous State -> COPY ALL edges/coefficients -> UPDATE only parts with new evidence
   - **PRESERVE known coefficients**: If a coefficient was determined in previous steps, DO NOT reset it to null unless contradicted
   - Update rule: MAY update when new evidence gathered; keep unchanged during reactor operations or when no new discovery
   - **CRITICAL**: Update your frequency equation by REASONING about causal relationships, NOT by curve-fitting past_data. Do NOT directly fit observation data. Before operating the reactor, your hypothesis MUST contain a definite frequency equation with all coefficients determined through causal reasoning.

5. **experiment** (JSON object): Specification of the planned intervention when using the Property Manipulator, example:
   - `{"target_prop":"PropName","target_value":number}`
   - If no intervention is available or you are not planning one on this step, use `{}` instead of inventing a fake intervention

6. **Execution action fields** (one of the following):
   - Non-dialog actions: `"action":"ACTION_NAME", "arg1":value, "arg2":value` (arg1/arg2 optional depending on action)
   - Dialog option selection: `"chosen_dialog_option_int":integer`
   - Dialog Value input: `"value":number`

**Output Format** (keys in exact order):
```json
{
  "thought": "string explanation",
  "past_data": [array of observation objects],
  "hypothesis": {hypothesis object},
  "experiment": {experiment specification},
  ... action execution fields ...
}
```

**Important Guidelines:**

Causal Discovery Methodology (CRITICAL):
- **Derive frequency equations through CAUSAL REASONING, not data fitting**: Your hypothesis should be built by understanding which properties causally influence frequency, NOT by curve-fitting past_data or observation data.
- **Use intervention experiments when available**: When you intervene on property A and observe changes in property B, this reveals causal relationships. If no interventions are available, extract as much structure as possible from the passive observations and be explicit about what remains ambiguous.
- **Before operating the reactor**: Your hypothesis MUST contain a complete frequency equation with all coefficients determined. Do not leave coefficients as null when transitioning to reactor operations.
- **For reactor frequency calculation**: Read crystal properties directly from the reactor dialog text (`dialog_box`).
- **One-way phase transition**: Once you open/talk to the Crystal Reactor, the Property Manipulator becomes locked and cannot be used again.
- **If a deterministic candidate-graph list is provided in Previous State**: treat it as a hard consistency filter derived from the configured graph family and the collected data. Use it to narrow your reasoning, but still explain why the remaining candidates differ.

Data Management:
- `past_data`: Always maintain complete history. When adding passive observations or intervention results, append to the array (don't replace). During reactor operations, keep past_data unchanged unless genuinely new evidence appears.
- `hypothesis`: Update edges when causal relationships are discovered or refuted. Update coefficients through causal reasoning from the full evidence set. Keep hypothesis frozen during reactor operations.
- `thought`: Must explain the logic connecting your current hypothesis/data to the next experiment. If previous actions failed, explain why and justify why the new action will succeed.
- `experiment`: Must precisely specify what you plan to do next and align with the execution action fields.

**Example of Incremental Update:**
```
Previous State shows:
  Last past_data: [{"id":"0/3","props":{"pH":50},"freq":100}]
  Last hypothesis: {"edges":[{"from":"pH","to":"resonanceFreq"}],"freq_equation":"resonanceFreq = base + c_pH*pH","coefficients":{"base":50,"c_pH":1}}

After new observation (pH=80, freq=130):
[CORRECT] CORRECT: {"past_data":[{"id":"0/3","props":{"pH":50},"freq":100},{"id":"1/3","props":{"pH":80},"freq":130}],"hypothesis":{"edges":[{"from":"pH","to":"resonanceFreq"}],"freq_equation":"resonanceFreq = base + c_pH*pH","coefficients":{"base":50,"c_pH":1}}}
[WRONG] WRONG: {"past_data":[{"id":"1/3","props":{"pH":80},"freq":130}],...}  // Missing previous entry!
[WRONG] WRONG: {"hypothesis":{"edges":[],"coefficients":{"base":null,...}}}  // Reset known coefficients!
[WRONG] WRONG: {"freq_equation":"resonanceFreq = base + c1*pH",...}  // Wrong coefficient naming! Should be c_pH
[WRONG] WRONG: {"coefficients":{"base":50,"c_pH":1,"c_pressure":0}}  // Don't include zero coefficients!
```

Hints:
- **CAUSAL REASONING FIRST**: Build your frequency equation by reasoning about causal relationships discovered through interventions, NOT by fitting curves to past_data. Your equation should reflect the causal structure, not just correlations.
- **REACTOR DIALOG FIRST**: When calculating target frequency, use the crystal properties shown directly in the reactor dialog.
- **DO NOT GO BACK**: After entering the reactor phase, do not return to the Property Manipulator. Continue with reactor frequency setting and activation.
- If your last action failed, or other last recent actions failed, please consider thinking why they failed, and trying different actions unless you believe things have changed to make failed actions work this time.
- If you don't see what you're looking for, and anticipate it might be in another location, consider teleporting to that location.
- Use verbose "thought" statements to maintain your progress, running hypotheses, your knowledge about the world, etc. Always include an ACCESSIBILITY CHECK: confirm any UUIDs you reference are from the interactable list/inventory now; otherwise first act to make them accessible.
- REMEMBER, IF YOU'RE GOING TO AN OBJECT, INSTEAD OF MOVING NORTH/EAST/SOUTH/WEST, or ROTATING, YOU SHOULD TRY TELEPORTING DIRECTLY TO OBJECTS.  IT'S MUCH FASTER AND LESS ERROR-PRONE. YOU CAN TELEPORT TO AN OBJECT NO MATTER WHERE IT IS, EVEN IF ITS NOT LISTED IN THE ACCESSIBLE OBJECTS LIST. JUST REMEBER ITS UUID.
- WHEN TELEPORTING, REMEMBER TO LOOK AT THE ENTIRE OBJECT LIST IN THE OBSERVATION, NOT JUST THE ACCESSIBLE OBJECTS LIST.
- FOR ACTIONS INVOLVING TWO ARGS (E.G. USE AND PUT), ONE OF THE OBJECTS MUST BE IN YOUR INVENTORY AND YOU MUST BE NEXT TO THE OTHER OBJECT. E.G. TO USE SHOVEL ON THE SOIL, YOU MUST HAVE THE SHOVEL IN YOUR INVENTORY AND BE NEXT TO THE SOIL.
- IF YOU FIND YOURSELF WAITING FOR SOMETHING TO HAPPEN, CHECK TO SEE IF IT'S ALREADY HAPPENED.  YOU MIGHT BE WAITING FOR SOMETHING THAT'S ALREADY HAPPENED.
- IF YOU FIND YOURSELF WAITING FOR A LONG TIME, YOU MIGHT BE STUCK. TRY TO REEXAMINE YOUR TASK, REASSESS WHERE YOU ARE, AND MAKE A NEW PLAN.
- IF YOU THINK YOU ARE DONE, SUBMIT WITH THE ACTION: {"action": "SUBMIT", "arg1": "Task completed!"}.
- IF YOU ARE NOT MAKING ANY PROGRESS AND CAN NOT FIGURE OUT HOW TO PROCEED, USE THE SUBMIT ACTION WITH ARG1 EXPLAINING YOUR REASONING, e.g., {"action": "SUBMIT", "arg1": "Task failed! Unable to explore the building!"}.
- If the target item is nearby but inaccessible, consider ROTATE_DIRECTION to help yourself face the target item before attempting to move or interact.

Task: {{ input_str }}

Current Environment Observation:
```json
{{ observation }}
```

{% if in_dialog %}
NOTE: You are currently in a dialog.
For reference, here is the dialog that you are currently in:
```json
{{ dialog_box }}
```

{% else %}
Valid Actions:
```json
{{ known_actions}}
```

Valid Teleport Locations:
```json
{{ teleport_destinations }}
```

CRITICAL accessibility rules (strict):
- For any `arg1`/`arg2`, choose strictly from the List of objects that are interactable (from your inventory, directly in front of you, or on adjacent tiles north/east/south/west): {{ interactable_objects }}. If an object is not in those lists, first make it accessible (MOVE/ROTATE/TELEPORT/PICK_UP) before acting.
- In your "thought", explicitly state which list you took each UUID from (inventory vs interactable list) and why it is accessible now. If not accessible, state the plan to make it accessible first.

List of objects that are interactable (from your inventory, and directly in front of you):
{{ interactable_objects }}

{% endif %}

Return ONLY a single JSON object. Keys MUST appear in this exact order: memory -> thought -> past_data -> hypothesis -> experiment -> (action+args OR chosen_dialog_option_int OR value). Do not include any extra text outside the JSON.

\end{tcblisting}

\begin{tcblisting}{
  promptbox,
  title={Phase 2 Reactor Task Prompt},
  listing only,
  listing options={
    basicstyle=\ttfamily\scriptsize,
    breaklines=true,
    columns=fullflexible,
    keepspaces=true,
    showstringspaces=false
  }
}
You are at the Causal Discovery Lab on Planet X. Quantum Crystals have interesting properties that may be causally related. 

=== YOUR MISSION ===

THEORETICAL BACKGROUND:

1. CRYSTAL RESONANCE FREQUENCY:
   - Frequency is determined by certain properties (range: 0-100) through LINEAR relationships
   - Frequency CANNOT be directly modified by human intervention
   - Frequency can only change indirectly as a causal consequence of other property changes

2. CAUSAL STRUCTURE:
   - Properties have LINEAR causal relationships with each other
   - These relationships form a DAG (Directed Acyclic Graph):
     * No bidirectional influences (if A affects B, B cannot affect A)
     * No cycles (no circular chains like A -> B -> C -> A)
     * Acyclic structures are allowed (e.g., A -> B, A -> C, B -> D, C -> D)

3. PROPERTY INTERVENTION:
   - You can intervene on ONE property at a time using the Property Manipulator
   - When you modify a property (e.g., prop_A):
     * Other properties CANNOT be simultaneously modified by direct intervention
     * However, other properties MAY change automatically as a causal consequence
     * Frequency MAY also change if it depends on the modified property or its effects

4. MATHEMATICAL MODEL:
   - Each property has a BASE VALUE that can be adjusted (except frequency)
   - Each property's value = base_value + sum of causal influences from other properties
   - Example: If frequency depends on properties A and B:
     frequency = freq_base_value + k_A * A + k_B * B
     (Note: freq_base_value is fixed and cannot be adjusted)
   - Example: If temperature depends on properties C, D, and E:
     temperature = temp_base_value + k_C * C + k_D * D + k_E * E
     (Note: When you adjust temperature, you are actually adjusting temp_base_value)
   - IMPORTANT: When you modify a property(frequency cannot be directly modified), you are changing its base_value,
     which then propagates through the causal graph according to the linear relationships

CAUSAL DISCOVERY STAGE (If the Property Manipulator is available)
- There is an EXPERIMENT CRYSTAL (quantum crystal 3) currently in the Property Manipulator
- Depending on the experiment setting, you may start with zero or more existing observations and zero or more remaining property adjustments
- If `{{ max_uses }}` is greater than 0, you may use the Property Manipulator to discover causal relationships
- If `{{ max_uses }}` is 0, rely on the available observations only and proceed without Property Manipulator interventions
- Goal: Figure out the relationship between controllable and derived properties as far as the available evidence allows

REACTOR ACTIVATION STAGE
- Once you have extracted all useful information from the available observations/interventions, activate the Crystal Reactor
- Calculate what the crystal's resonance frequency should be based on its properties
- Place the target crystal into the Crystal Reactor
- Set the reactor's frequency to match the crystal's resonance frequency
- Tolerance: +/-{{ frequency_tolerance }} Hz

IMPORTANT NOTICE:
- The EXPERIMENT CRYSTAL (used in Phase 1) and the TARGET CRYSTAL (used in Phase 2) are DIFFERENT crystals
- They are not movable between locations
- You CANNOT directly measure the target crystal's frequency - you must calculate it from its properties
- The Property Manipulator shows current crystal properties
- The Property Manipulator has a usage counter showing remaining adjustments

EXISTING OBSERVATIONS (possibly zero, from previous experiments):
{{ existing_observations }}

IMPORTANT USAGE INSTRUCTIONS:
- Property Manipulator: Use TALK action, e.g. { "action": "TALK", "arg1": <property_manipulator_uuid> }
  - After dialog opens, the main menu shows available properties to adjust (e.g., Temperature, Moisture, etc.)
  - Step 1: Select the property you want to adjust, e.g., {"chosen_dialog_option_int": 1} for the first property
  - Step 2: You will enter the value input mode DIRECTLY (no intermediate menu)
  - Step 3: In value input mode, provide ONLY the "value" field in your JSON - do NOT select dialog options
  - Example workflow:
    * First TALK: {"reasoning": "Selecting property to adjust", "past_data": [...], "hypothesis": {...}, "experiment": {...}, "action": "TALK", "arg1": <uuid>, "chosen_dialog_option_int": 1}
    * Next action: {"reasoning": "Setting property to test hypothesis", "past_data": [...], "hypothesis": {...}, "experiment": {"target_prop":"PropName","target_value":50,"purpose":"..."}, "value": 50}  # Set value (NOT chosen_dialog_option_int!)
  - CRITICAL: When in value input mode, you will see "[Do not select this option]" in dialog options
    * This is NOT a real option - ignore it!
    * Simply provide the "value" field with your desired number
    * After setting the value, you automatically return to the main menu
  - IMPORTANT: Each property adjustment counts toward your usage limit when interventions are available.
  - Watch the usage counter if it is positive; in some experiment settings it may already be zero.
  - CRITICAL: Once you use the Crystal Reactor (set its frequency), you CANNOT use the Property Manipulator anymore!
  - Plan your experiments carefully: use Property Manipulator FIRST, then activate the reactor.

- Crystal Reactor: Use TALK action, e.g. { "action": "TALK", "arg1": <crystal_reactor_uuid> }
  - This opens a dialog menu showing the current resonance frequency (range: 0-10,000 Hz)
  - Step 1: Select option 1 ("Set frequency to a specific value") to enter frequency input mode
  - Step 2: After "is_in_dialog" is True and you are in frequency input mode, include the numeric "value" field in your JSON with the desired value (0-10000 Hz)
  - Example: {..., "chosen_dialog_option_int": 1} to select input mode, then {..., "value": <frequency_value>} to set frequency
  - You can select "Back to main menu" to return to the root menu
  - The reactor activates when: (1) target crystal is inside, AND (2) frequency matches crystal's frequency

\end{tcblisting}

\subsection{DSL Implementation Details}
\label{app:dsl_implementation}

The DSL hypothesis schema is presented with field-by-field descriptions and a
small worked example so the model knows the expected key names, edge format,
coefficient format, and convention for referring to \texttt{frequency}. The
episode configuration fixes the property names and functional family, so the
parser only accepts variables emitted by the environment prompt. If a record
fails the schema check, for example because it contains an undeclared property,
a non-numeric coefficient, or a mismatch between listed parents and equation
terms, we re-prompt up to two times with the parser error before recording a
parse failure for that step.

\subsection{Trajectory-Level DSL Visualization}
\label{app:dsl_visualization}

\begin{figure*}[t]
\centering
\begin{minipage}[t]{0.49\textwidth}
\centering
\includegraphics[width=\linewidth]{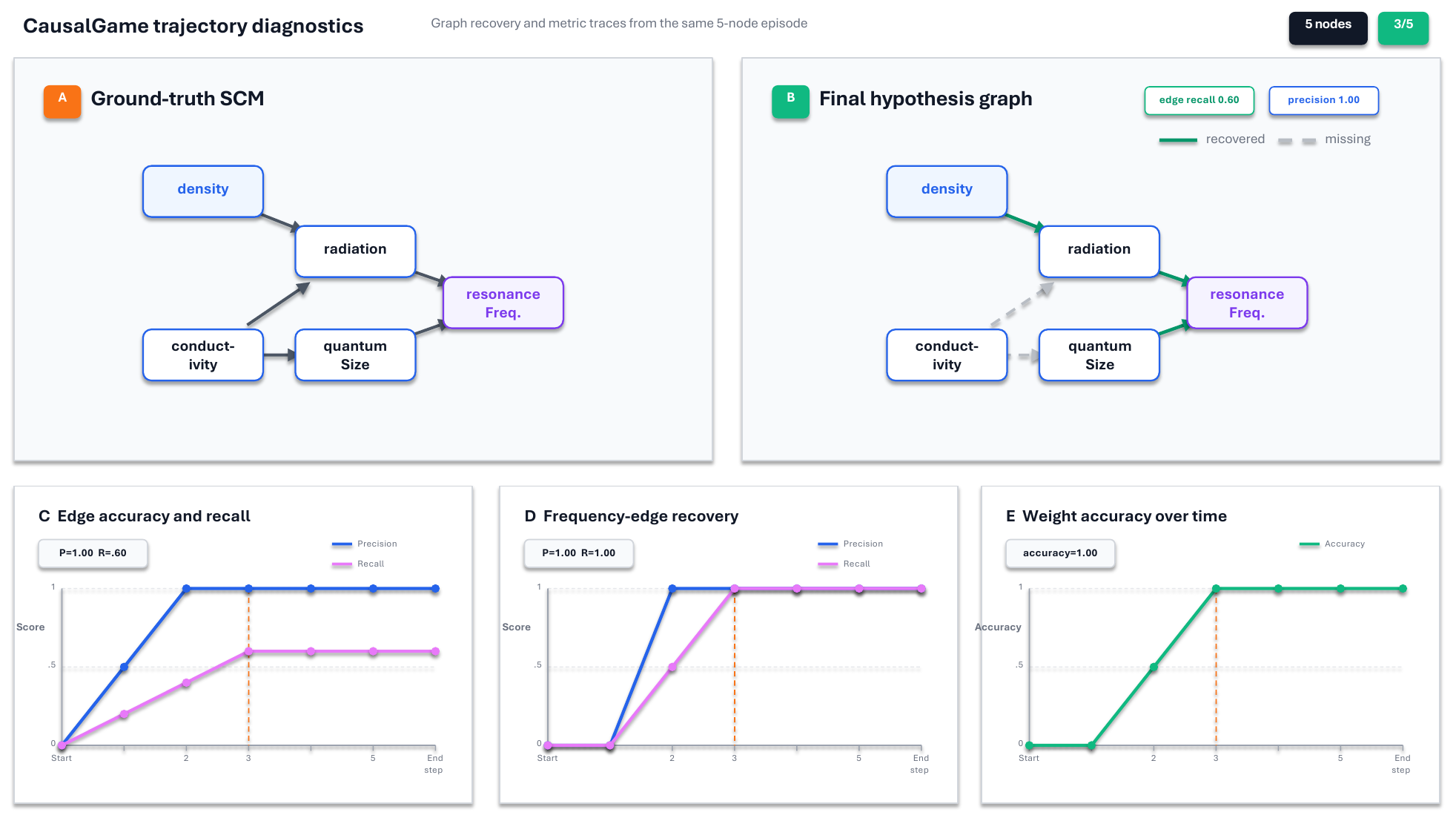}\\
{\small (a) Paper-ready trajectory schematic.}
\end{minipage}\hfill
\begin{minipage}[t]{0.49\textwidth}
\centering
\includegraphics[width=\linewidth]{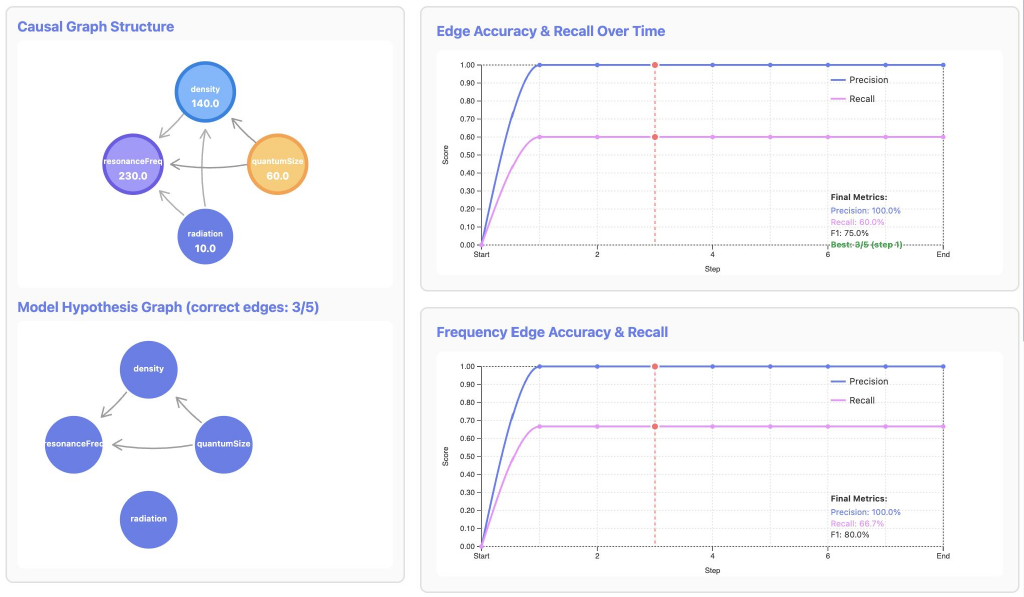}\\
{\small (b) Visualization-platform screenshot.}
\end{minipage}
\caption{Trajectory-level causal graph visualizations in \projectname{}. The schematic and screenshot both expose the ground-truth graph, the agent's hypothesis graph, and recovery metrics over the intervention sequence.}
\label{fig:causal_visualization}
\end{figure*}

The DSL records make the agent's evolving causal hypothesis auditable, rather
than leaving mechanism recovery to be inferred from a final reactor prediction.
Figure~\ref{fig:causal_visualization} is included because the DSL and task
sections refer to this trajectory-level view: it shows the ground-truth graph,
the agent hypothesis graph, and recovery metrics over the interaction sequence.

\subsection{Mechanism Robustness and Perturbation Controls}
\label{app:robustness_controls}

\begin{figure}[!htbp]
\centering
\includegraphics[width=0.70\linewidth]{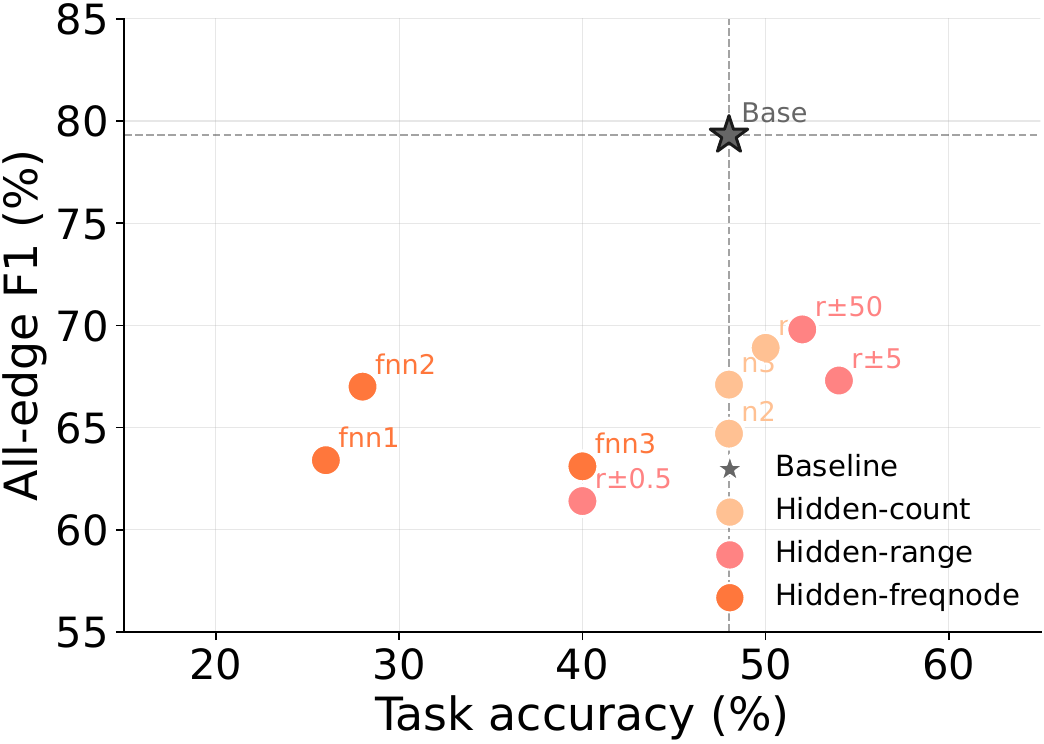}
\caption{Hidden-noise diagnostics across all ten 4-node settings, plotted as
(task accuracy, all-edge $F_1$) and colored by noise category. The dashed
cross marks the unperturbed baseline. Hidden-count and Hidden-range settings
drop $F_1$ but keep accuracy near baseline; the \texttt{freqnode} family
collapses accuracy without further $F_1$ loss, exposing a fragile
parent--target shortcut.}
\label{fig:hidden_variables}
\end{figure}

\begin{figure}[t]
\centering
\includegraphics[width=0.62\linewidth]{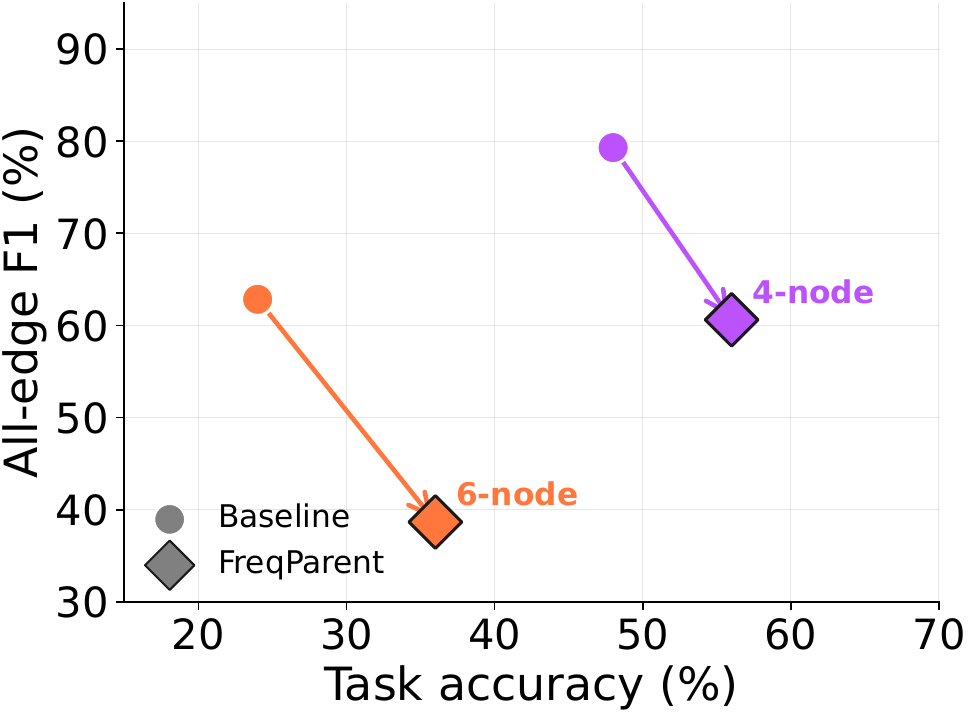}
\caption{\texttt{FreqParent} experiments on \texttt{GPT-5-mini} in
(task accuracy, all-edge $F_1$) space. Baseline $\to$ \texttt{FreqParent}
arrows move down-right on both graph sizes: prediction accuracy rises while
graph fidelity falls, especially on 6-node graphs.}
\label{fig:FreqParent_followup_metrics}
\end{figure}

The main text uses these results to separate task success from causal
faithfulness. Table~\ref{tab:linear_quad_metrics} supports the hard-quadratic
comparison by showing that the main degradation is in the recovered
\texttt{frequency} mechanism, not root discovery. Figure~\ref{fig:hidden_variables}
and Table~\ref{tab:hidden_metrics} expand the hidden-noise analysis: ordinary
hidden count/range perturbations mostly reduce graph \(F_1\), while perturbing
the \texttt{frequency} target family sharply reduces task accuracy. The
\texttt{FreqParent} follow-up in Figure~\ref{fig:FreqParent_followup_metrics}
and Table~\ref{tab:freqparent_followup_metrics} tests a related modelling
choice: allowing \texttt{frequency} outgoing edges improves prediction but
hurts full-graph fidelity.

\begin{table}[t]
\centering
\small
\setlength{\tabcolsep}{4pt}
\begin{tabular}{lccccc}
\toprule
Setting & Acc. & Root \(F_1\) & All-edge P/R/\(F_1\) & Freq-edge \(F_1\) & Freq-weight \(F_1\) \\
\midrule
GPT-5-mini 4-node linear reference & 48 & 0.559 & 0.833/0.778/0.793 & 0.812 & 0.589 \\
GPT-5-mini 4-node quad-hard & 24 & 0.829 & 0.902/0.667/0.736 & 0.741 & 0.251 \\
\bottomrule
\end{tabular}
\caption{Linear versus hard-quadratic mechanisms on matched 4-node graphs. The main loss under quadratic dynamics is not root discovery but identifying the correct frequency mechanism.}
\label{tab:linear_quad_metrics}
\end{table}

\begin{table}[!htbp]
\centering
\small
\setlength{\tabcolsep}{4pt}
\begin{tabular}{lccccc}
\toprule
Setting & Acc. & Root \(F_1\) & All-edge P/R/\(F_1\) & Freq-edge \(F_1\) & Freq-weight \(F_1\) \\
\midrule
4-node main baseline & 48 & 0.559 & 0.833/0.778/0.793 & 0.812 & 0.589 \\
\midrule
hidden count \(=1\) & 50 & 0.708 & 0.854/0.610/0.689 & 0.781 & 0.477 \\
hidden count \(=2\) & 48 & 0.747 & 0.838/0.576/0.647 & 0.781 & 0.510 \\
hidden count \(=3\) & 48 & 0.672 & 0.883/0.581/0.671 & 0.799 & 0.493 \\
hidden range \(=\pm 0.5\) & 40 & 0.666 & 0.818/0.534/0.614 & 0.729 & 0.389 \\
hidden range \(=\pm 5\) & 54 & 0.755 & 0.876/0.610/0.673 & 0.766 & 0.546 \\
hidden range \(=\pm 50\) & 52 & 0.675 & 0.920/0.605/0.698 & 0.818 & 0.567 \\
\midrule
hidden \texttt{freqnode} \(=1\) & 26 & 0.805 & 0.800/0.573/0.634 & 0.678 & 0.218 \\
hidden \texttt{freqnode} \(=2\) & 28 & 0.699 & 0.851/0.603/0.670 & 0.735 & 0.327 \\
hidden \texttt{freqnode} \(=3\) & 40 & 0.623 & 0.850/0.537/0.631 & 0.789 & 0.361 \\
\bottomrule
\end{tabular}
\caption{Hidden-noise robustness on 4-node graphs under exact-context re-evaluation. The table includes the 4-node main baseline, standard hidden settings, and controlled \texttt{hidden-freqnode} settings where \texttt{resonanceFreq} is explicitly included among hidden targets.}
\label{tab:hidden_metrics}
\end{table}

\begin{table}[!ht]
\centering
\small
\setlength{\tabcolsep}{4pt}
\begin{tabular}{llccccc}
\toprule
Suite & Setting & \(n\) & Acc. & All-edge \(F_1\) & Freq-edge \(F_1\) & Freq-weight \(F_1\) \\
\midrule
4-node main & Baseline & 50 & 48.0 & 0.793 & 0.812 & 0.589 \\
4-node main & FreqParent & 50 & 56.0 & 0.606 & 0.733 & 0.451 \\
6-node main & Baseline & 50 & 24.0 & 0.628 & 0.679 & 0.455 \\
6-node main & FreqParent & 50 & 36.0 & 0.387 & 0.582 & 0.364 \\
\bottomrule
\end{tabular}
\caption{FreqParent follow-up under exact-context re-evaluation (GPT-5-mini). Allowing \texttt{resonanceFreq} as a parent improves prediction but degrades full-graph recovery.}
\label{tab:freqparent_followup_metrics}
\end{table}

\subsection{Intervention-Trace Controls}
\label{app:intervention_trace_controls}

The Golden follow-up tests whether better intervention traces alone are enough
to recover a faithful SCM. Table~\ref{tab:golden_followup_metrics} supports
the main-text claim that low-MEC traces greatly improve endpoint prediction,
yet do not produce matching gains in all-edge recovery.

\begin{table}[!ht]
\centering
\small
\setlength{\tabcolsep}{4pt}
\begin{tabular}{llccccc}
\toprule
Suite & Setting & \(n\) & Acc. & All-edge \(F_1\) & Freq-edge \(F_1\) & Freq-weight \(F_1\) \\
\midrule
4-node main & Baseline & 50 & 48.0 & 0.793 & 0.812 & 0.589 \\
4-node main & Golden & 50 & 90.0 & 0.728 & 0.743 & 0.644 \\
6-node main & Baseline & 50 & 24.0 & 0.628 & 0.679 & 0.455 \\
6-node main & Golden & 50 & 44.0 & 0.574 & 0.715 & 0.498 \\
\bottomrule
\end{tabular}
\caption{Golden follow-up under exact-context re-evaluation (GPT-5-mini). Injected low-MEC intervention traces strongly improve frequency prediction but do not improve all-edge recovery.}
\label{tab:golden_followup_metrics}
\end{table}

\subsection{Model Family and Graph-Size Scaling}
\label{app:model_scale_results}
\label{app:full_experiment_tables}

\begin{figure}[t]
\centering
\begin{minipage}[t]{0.49\linewidth}
\centering
\includegraphics[width=\linewidth]{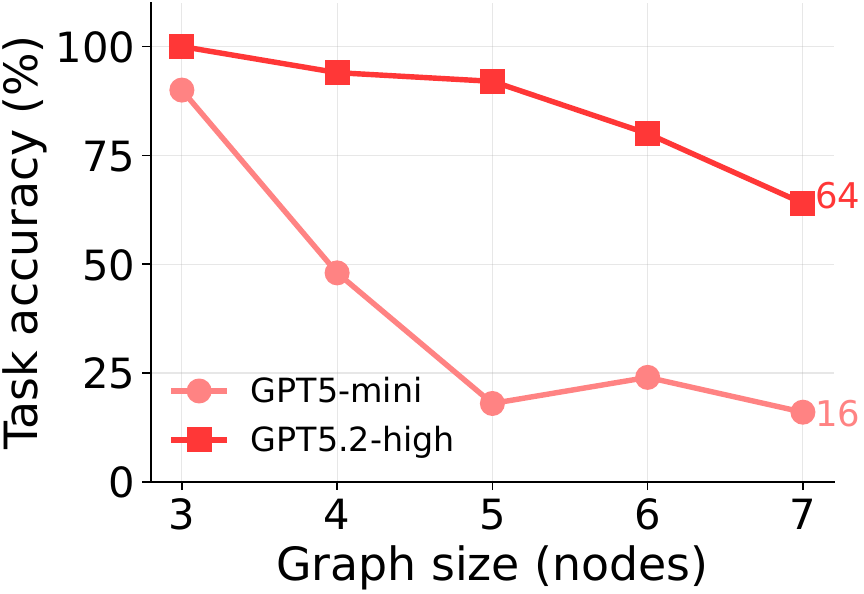}\\
{\small (a) Task accuracy.}
\end{minipage}\hfill
\begin{minipage}[t]{0.49\linewidth}
\centering
\includegraphics[width=\linewidth]{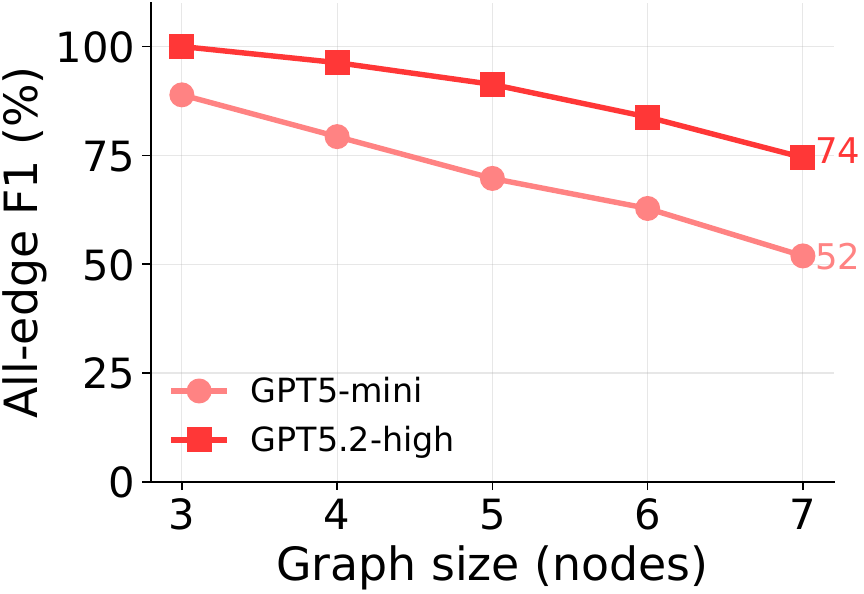}\\
{\small (b) All-edge $F_1$.}
\end{minipage}

\vspace{0.5em}

\begin{minipage}[t]{0.49\linewidth}
\centering
\includegraphics[width=\linewidth]{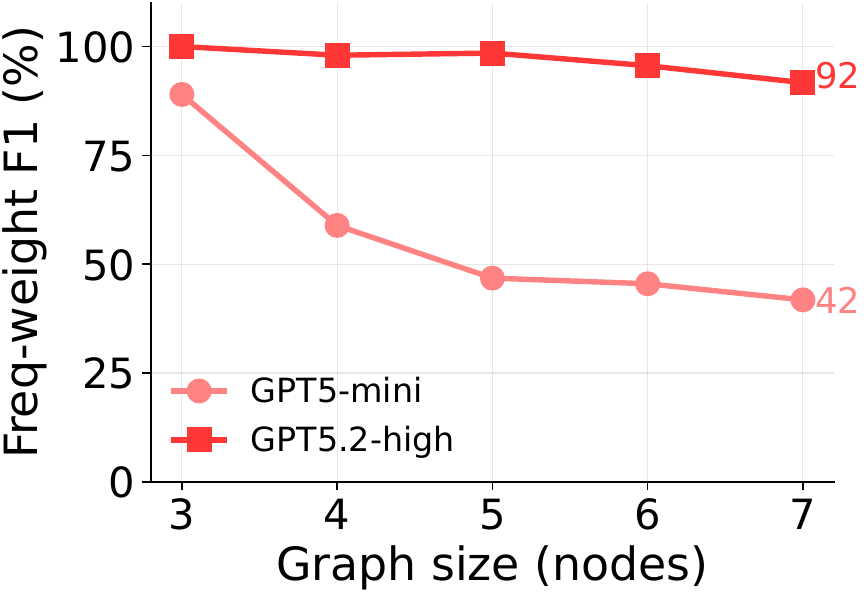}\\
{\small (c) \texttt{frequency}-weight $F_1$.}
\end{minipage}\hfill
\begin{minipage}[t]{0.49\linewidth}
\centering
\includegraphics[width=\linewidth]{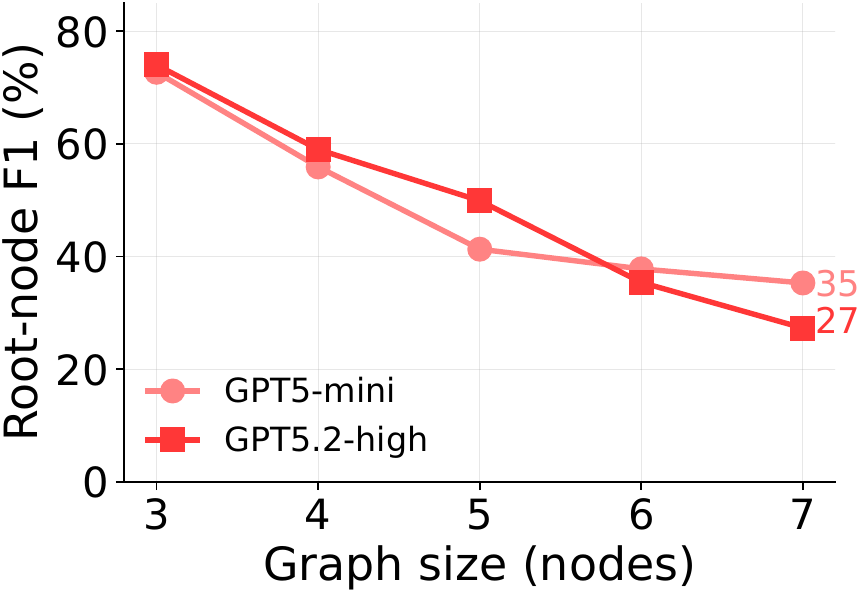}\\
{\small (d) Root-node $F_1$.}
\end{minipage}
\caption{Absolute trajectories of the four recovery metrics across graph
sizes for \texttt{GPT-5-mini} and \texttt{GPT-5.2-high}. Task accuracy and
\texttt{frequency}-weight $F_1$ separate the two models cleanly; root-node
$F_1$ shows the smallest gap and both models still drop on 6--7 node
graphs.}
\label{fig:scaling_metric_curves}
\end{figure}

These tables give the full numeric support for the model-family and
graph-size claims in RQ3. Table~\ref{tab:gpt_model_metrics} combines the
3--7 node GPT and Qwen sweeps behind the model-family radar plots, while
Figure~\ref{fig:scaling_metric_curves} shows the corresponding GPT absolute
metric trajectories (the capability-gap decomposition is in
Figure~\ref{fig:model_metric_support} in the main text).
Table~\ref{tab:simple_graph_imec} keeps the small-graph ambiguity diagnostic:
even when task accuracy is high, the final hypotheses do not necessarily
identify a singleton SCM.

\begin{table}[!ht]
\centering
\small
\setlength{\tabcolsep}{4pt}
\begin{tabular}{llccccccc}
\toprule
Model & Nodes & Rows & Acc. & Root \(F_1\) & All-edge P/R/\(F_1\) & SHD\(\downarrow\) & Freq-edge \(F_1\) & Freq-weight \(F_1\) \\
\midrule
\texttt{GPT-5-mini} & 3 & 50 & 90.00 & 0.727 & 0.907/0.880/0.889 & 0.458 & 0.907 & 0.890 \\
\texttt{GPT-5-mini} & 4 & 50 & 48.00 & 0.559 & 0.833/0.778/0.793 & 1.766 & 0.812 & 0.589 \\
\texttt{GPT-5-mini} & 5 & 50 & 18.00 & 0.413 & 0.825/0.629/0.697 & 3.543 & 0.687 & 0.468 \\
\texttt{GPT-5-mini} & 6 & 50 & 24.00 & 0.379 & 0.739/0.574/0.628 & 5.490 & 0.679 & 0.455 \\
\texttt{GPT-5-mini} & 7 & 50 & 16.00 & 0.353 & 0.700/0.449/0.519 & 7.511 & 0.585 & 0.418 \\
\midrule
\texttt{GPT-5.2-high} & 3 & 50 & 100.00 & 0.740 & 1.000/1.000/1.000 & 0.000 & 1.000 & 1.000 \\
\texttt{GPT-5.2-high} & 4 & 50 & 94.00 & 0.590 & 0.957/0.972/0.963 & 0.391 & 0.977 & 0.980 \\
\texttt{GPT-5.2-high} & 5 & 50 & 92.00 & 0.499 & 0.946/0.896/0.913 & 1.319 & 0.980 & 0.985 \\
\texttt{GPT-5.2-high} & 6 & 50 & 80.00 & 0.353 & 0.878/0.840/0.838 & 2.660 & 0.956 & 0.956 \\
\texttt{GPT-5.2-high} & 7 & 50 & 64.00 & 0.273 & 0.855/0.703/0.745 & 4.761 & 0.921 & 0.917 \\
\midrule
\texttt{Qwen3.5 non-thinking} & 3 & 50 & 72.00 & 0.713 & 0.840/0.717/0.761 & 1.060 & 0.833 & 0.870 \\
\texttt{Qwen3.5 non-thinking} & 4 & 50 & 52.00 & 0.526 & 0.648/0.492/0.539 & 3.520 & 0.640 & 0.691 \\
\texttt{Qwen3.5 non-thinking} & 5 & 50 & 22.00 & 0.551 & 0.727/0.506/0.570 & 4.840 & 0.612 & 0.513 \\
\texttt{Qwen3.5 non-thinking} & 6 & 50 & 20.00 & 0.341 & 0.604/0.374/0.430 & 7.620 & 0.575 & 0.493 \\
\texttt{Qwen3.5 non-thinking} & 7 & 50 & 22.00 & 0.370 & 0.483/0.284/0.331 & 10.800 & 0.497 & 0.508 \\
\midrule
\texttt{Qwen3.5 thinking} & 3 & 49 & 95.92 & 0.823 & 0.884/0.745/0.796 & 0.898 & 0.884 & 0.983 \\
\texttt{Qwen3.5 thinking} & 4 & 50 & 52.00 & 0.663 & 0.711/0.559/0.599 & 3.180 & 0.680 & 0.739 \\
\texttt{Qwen3.5 thinking} & 5 & 50 & 36.00 & 0.586 & 0.760/0.503/0.578 & 5.040 & 0.717 & 0.614 \\
\texttt{Qwen3.5 thinking} & 6 & 50 & 34.00 & 0.412 & 0.706/0.403/0.475 & 6.620 & 0.704 & 0.572 \\
\texttt{Qwen3.5 thinking} & 7 & 50 & 36.00 & 0.463 & 0.609/0.323/0.388 & 9.080 & 0.630 & 0.568 \\
\bottomrule
\end{tabular}
\caption{GPT and Qwen model comparison across the 3--7 node main suites. Acc.\ is endpoint reactor accuracy in percent; each P/R/\(F_1\) cell reports precision, recall, and \(F_1\) for the corresponding recovery target. SHD is directed all-edge structural Hamming distance, with reversed edges counted as one error.}
\label{tab:gpt_model_metrics}
\label{tab:qwen_model_metrics}
\end{table}

\begin{table}[!ht]
\centering
\small
\setlength{\tabcolsep}{4pt}
\begin{tabular}{lcc}
\toprule
Model & 3-node IMEC & 4-node IMEC \\
\midrule
GPT-5-mini & 1.920 & 4.540 \\
GPT-5.2-high & 1.480 & 3.620 \\
\bottomrule
\end{tabular}
\caption{Final IMEC on the simple 3- and 4-node main settings.}
\label{tab:simple_graph_imec}
\end{table}

Even on simple 3--4 node settings, high task success does not imply sufficient
causal disambiguation: the models' own interventions leave final IMEC above
one, so they do not reduce the possible SCM set to a singleton.

\subsection{Observation--Intervention Scaling}
\label{app:observation_intervention_scaling}

\begin{figure}[!htbp]
\centering
\begin{minipage}[t]{0.49\linewidth}
\centering
\includegraphics[width=\linewidth]{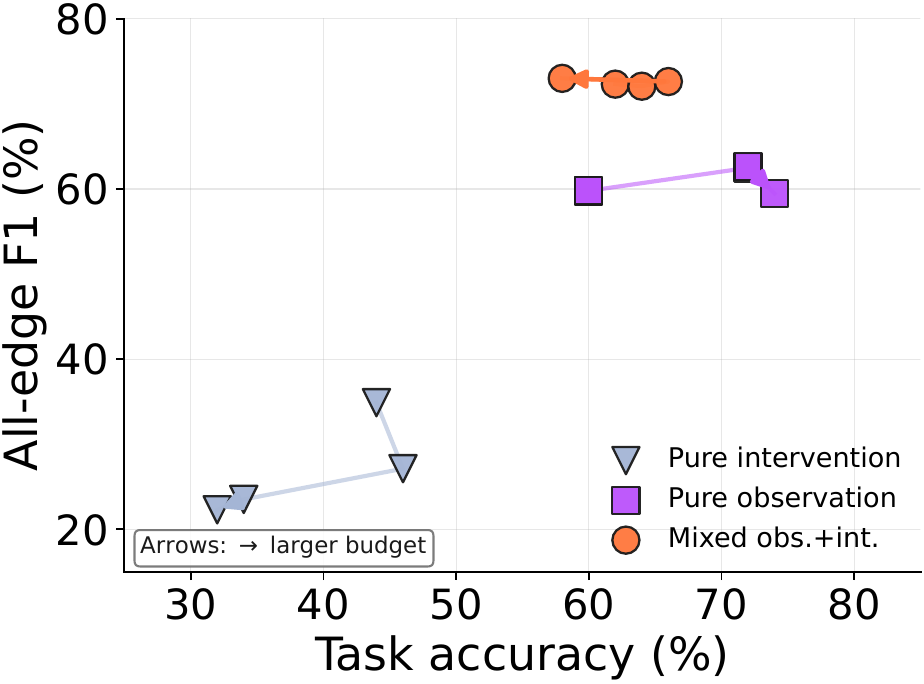}\\
{\small (a) \texttt{GPT-5-mini}, 4 nodes.}
\end{minipage}\hfill
\begin{minipage}[t]{0.49\linewidth}
\centering
\includegraphics[width=\linewidth]{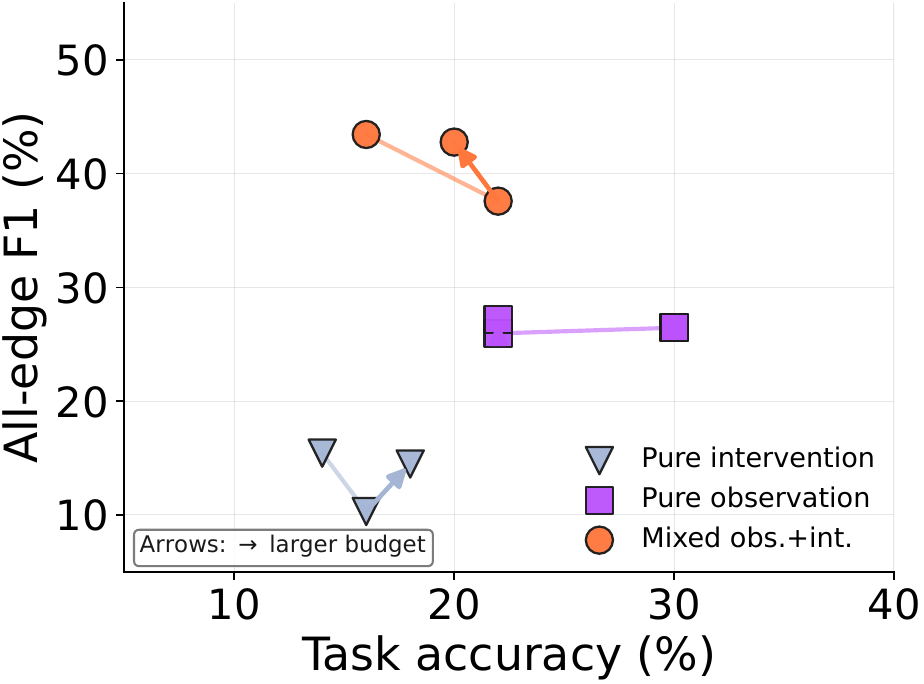}\\
{\small (b) \texttt{GPT-5-mini}, 6 nodes.}
\end{minipage}

\vspace{0.5em}

\begin{minipage}[t]{0.49\linewidth}
\centering
\includegraphics[width=\linewidth]{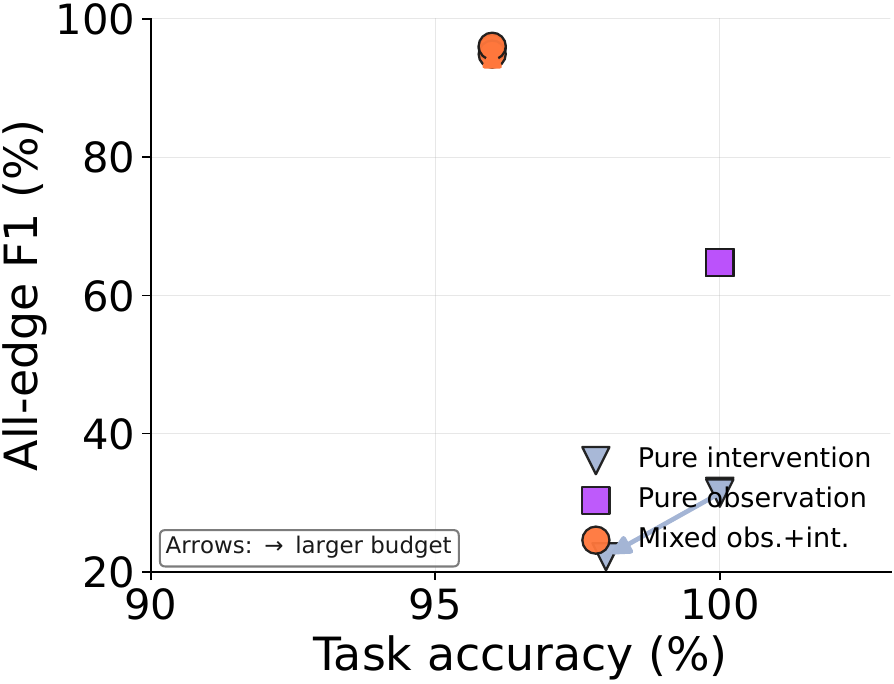}\\
{\small (c) \texttt{GPT-5.2-high}, 4 nodes.}
\end{minipage}\hfill
\begin{minipage}[t]{0.49\linewidth}
\centering
\includegraphics[width=\linewidth]{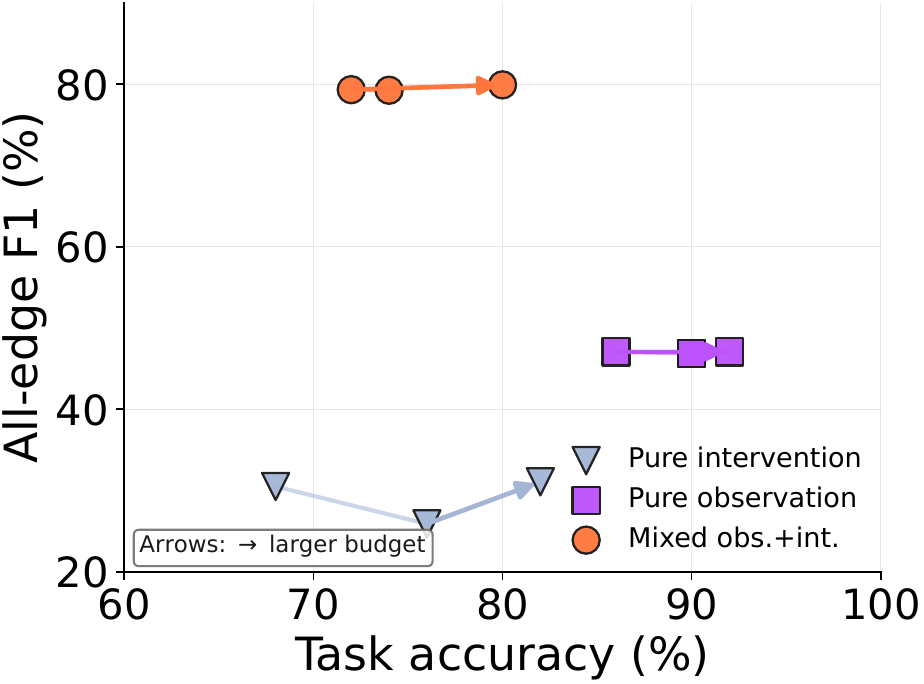}\\
{\small (d) \texttt{GPT-5.2-high}, 6 nodes.}
\end{minipage}
\caption{All scaling-suite settings plotted as (task accuracy, all-edge $F_1$),
colored by regime. Each regime is drawn as a polyline through its settings
in order of budget; the arrow on the final segment points toward the
largest budget. Mixed obs.+int.\ reaches the upper part of the
frontier; pure observation buys task accuracy without lifting
$F_1$; pure intervention stays near the lower-left. The pattern is
consistent across both models and both graph sizes.}
\label{fig:regime_scatter}
\end{figure}

\begin{figure}[!htbp]
\centering
\begin{minipage}[t]{0.485\linewidth}
\centering
\includegraphics[width=\linewidth]{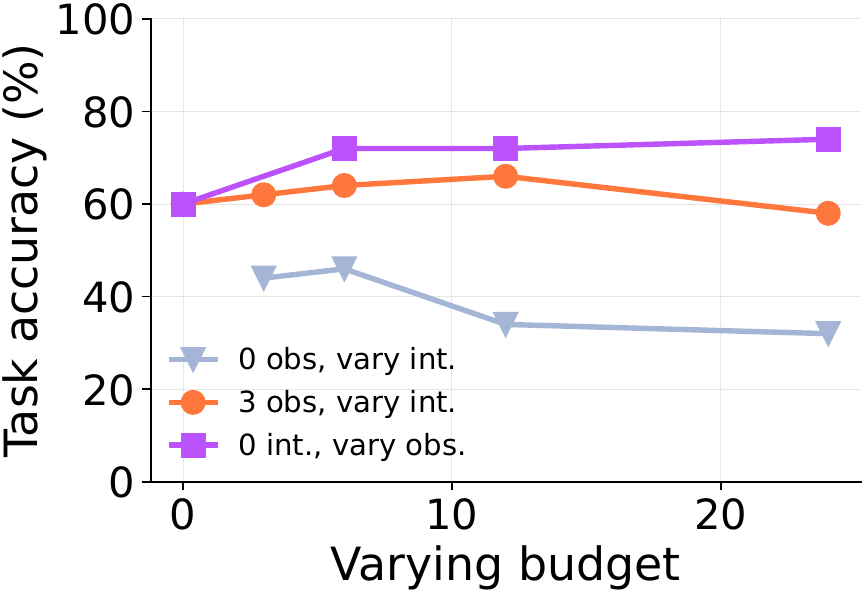}

\vspace{0.3em}
{\small GPT-5-mini, 4 nodes.}
\end{minipage}\hfill
\begin{minipage}[t]{0.485\linewidth}
\centering
\includegraphics[width=\linewidth]{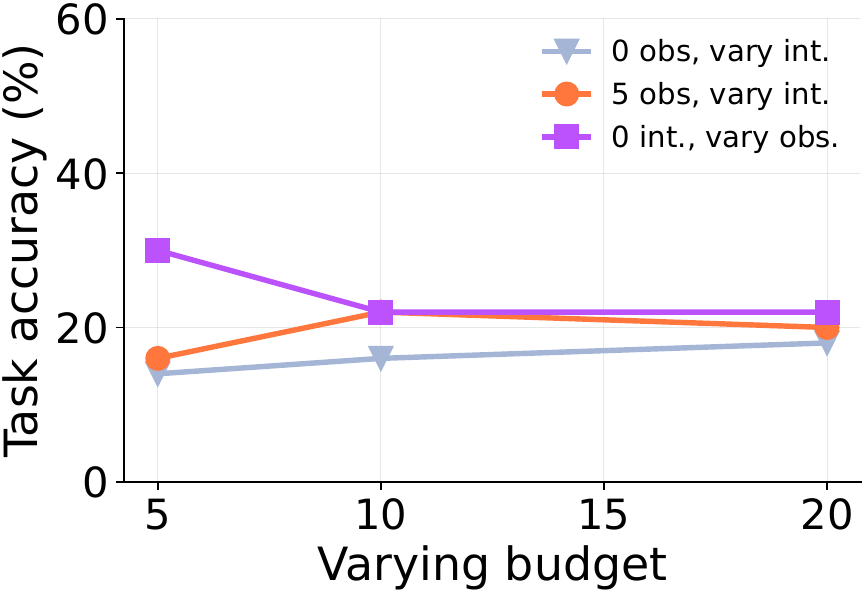}

\vspace{0.3em}
{\small GPT-5-mini, 6 nodes.}
\end{minipage}

\vspace{0.5em}

\begin{minipage}[t]{0.485\linewidth}
\centering
\includegraphics[width=\linewidth]{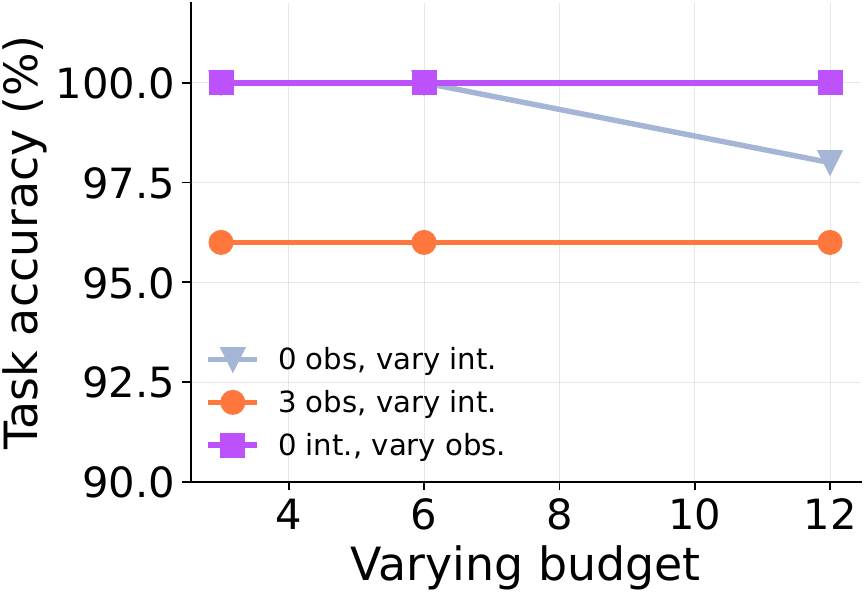}

\vspace{0.3em}
{\small GPT-5.2-high, 4 nodes.}
\end{minipage}\hfill
\begin{minipage}[t]{0.485\linewidth}
\centering
\includegraphics[width=\linewidth]{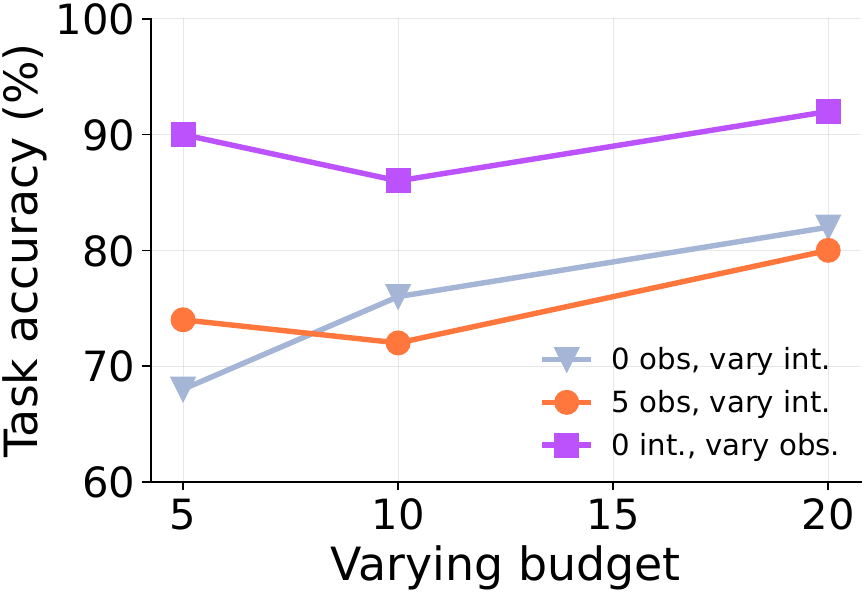}

\vspace{0.3em}
{\small GPT-5.2-high, 6 nodes.}
\end{minipage}
\caption{Observation/intervention scaling curves of \texttt{frequency} prediction score under different interaction modes.}
\label{fig:app_scaling_curves}
\end{figure}

\begin{figure}[!htbp]
\centering
\begin{minipage}[t]{0.485\linewidth}
\centering
\includegraphics[width=\linewidth]{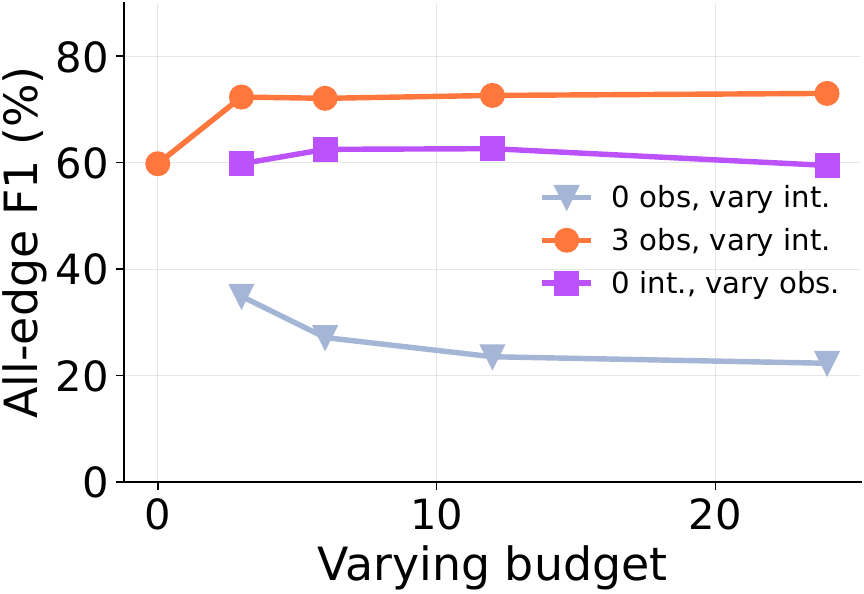}

\vspace{0.3em}
{\small GPT-5-mini, 4 nodes.}
\end{minipage}\hfill
\begin{minipage}[t]{0.485\linewidth}
\centering
\includegraphics[width=\linewidth]{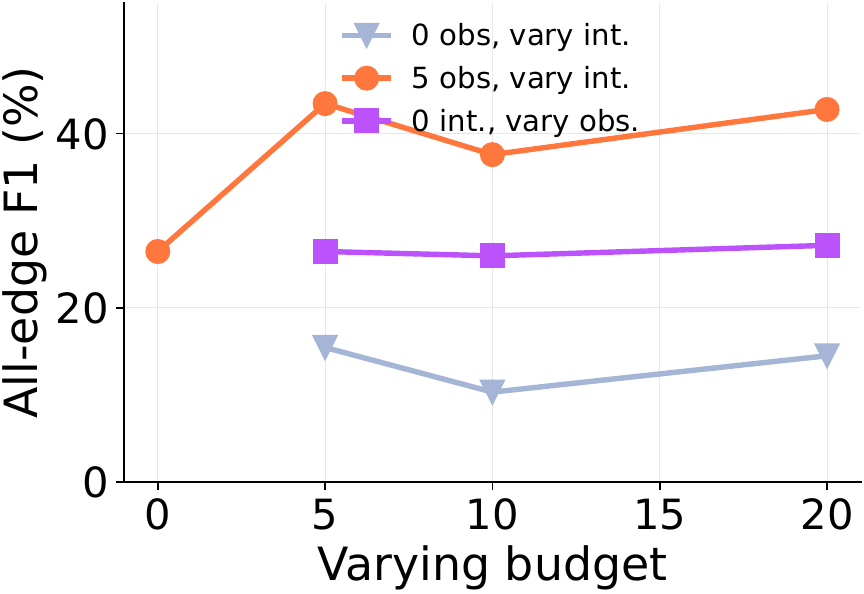}

\vspace{0.3em}
{\small GPT-5-mini, 6 nodes.}
\end{minipage}

\vspace{0.5em}

\begin{minipage}[t]{0.485\linewidth}
\centering
\includegraphics[width=\linewidth]{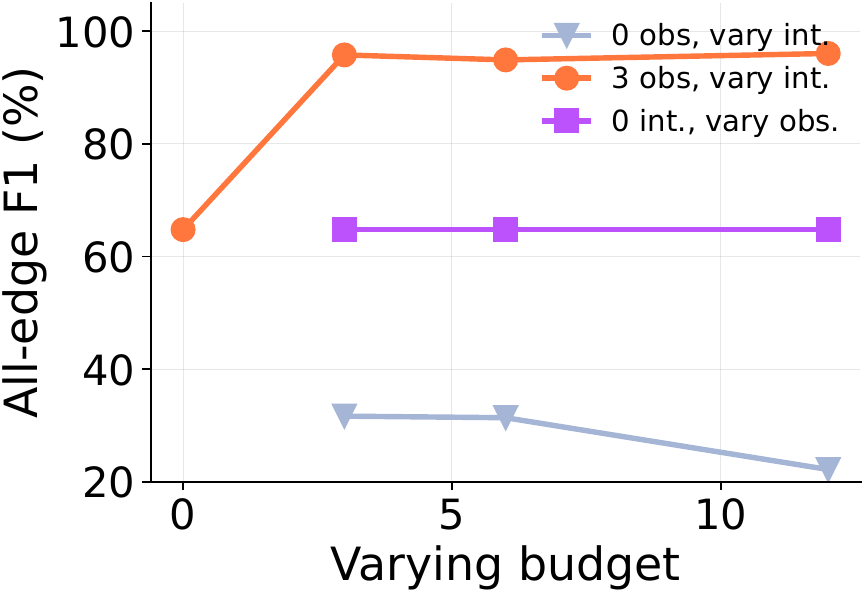}

\vspace{0.3em}
{\small GPT-5.2-high, 4 nodes.}
\end{minipage}\hfill
\begin{minipage}[t]{0.485\linewidth}
\centering
\includegraphics[width=\linewidth]{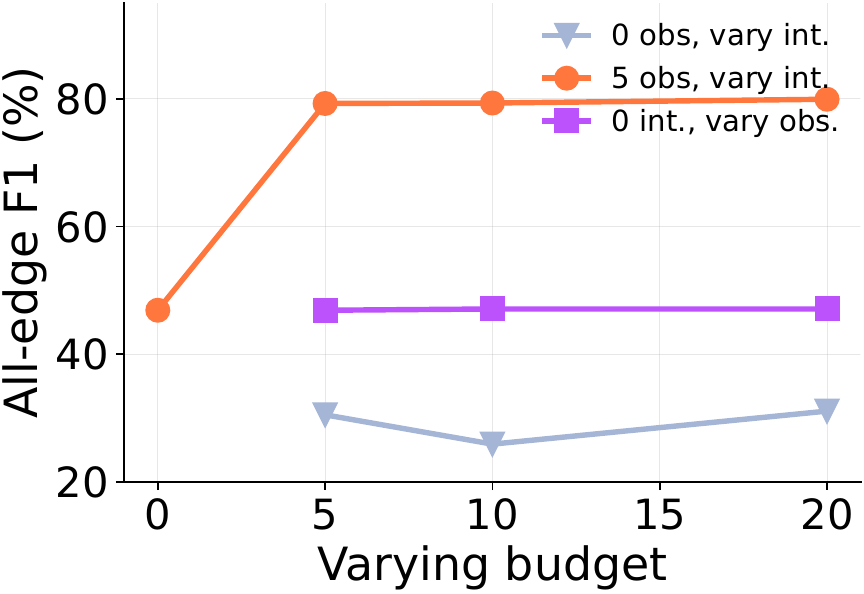}

\vspace{0.3em}
{\small GPT-5.2-high, 6 nodes.}
\end{minipage}
\caption{Observation/intervention scaling curves of all-edge recovery $F_1$ score under different interaction modes.}
\label{fig:app_reeval_scaling}
\end{figure}

The scaling appendix supports the RQ2 claim that observations and interventions
help different axes of performance. Figure~\ref{fig:regime_scatter} summarizes
the regime-level frontier, Figures~\ref{fig:app_scaling_curves}
and~\ref{fig:app_reeval_scaling} show the task and all-edge recovery curves,
and Tables~\ref{tab:scaling_mini_4nodes_metrics}--\ref{tab:scaling_gpt52_6nodes_metrics}
provide the numeric metric breakdowns for each model/size suite.

The next two tables unpack the \texttt{GPT-5-mini} scaling runs behind the
frontier plot. They show the same asymmetry as Figure~\ref{fig:regime_scatter}:
observation-only budgets often improve endpoint prediction, while mixed
observation--intervention budgets are the ones that consistently lift explicit
graph recovery.

\begin{table}[!ht]
\centering
\small
\setlength{\tabcolsep}{4pt}
\begin{tabular}{lccccc}
\toprule
Setting & Acc. & Root \(F_1\) & All-edge P/R/\(F_1\) & Freq-edge \(F_1\) & Freq-weight \(F_1\) \\
\midrule
\multicolumn{6}{l}{\textit{Pure observation}} \\
\(3\mathrm{o}0\mathrm{i}\) & 60 & 0.542 & 0.905/0.461/0.598 & 0.869 & 0.610 \\
\(6\mathrm{o}0\mathrm{i}\) & 72 & 0.545 & 0.930/0.486/0.625 & 0.901 & 0.727 \\
\(12\mathrm{o}0\mathrm{i}\) & 72 & 0.572 & 0.913/0.489/0.626 & 0.861 & 0.719 \\
\(24\mathrm{o}0\mathrm{i}\) & 74 & 0.542 & 0.898/0.460/0.595 & 0.864 & 0.740 \\
\midrule
\multicolumn{6}{l}{\textit{Pure intervention}} \\
\(0\mathrm{o}3\mathrm{i}\) & 44 & 0.315 & 0.416/0.318/0.349 & 0.364 & 0.341 \\
\(0\mathrm{o}6\mathrm{i}\) & 46 & 0.243 & 0.349/0.234/0.271 & 0.318 & 0.315 \\
\(0\mathrm{o}12\mathrm{i}\) & 34 & 0.208 & 0.347/0.194/0.235 & 0.277 & 0.185 \\
\(0\mathrm{o}24\mathrm{i}\) & 32 & 0.213 & 0.300/0.196/0.223 & 0.251 & 0.239 \\
\midrule
\multicolumn{6}{l}{\textit{Mixed intervention}} \\
\(3\mathrm{o}0\mathrm{i}\) & 60 & 0.542 & 0.905/0.461/0.598 & 0.869 & 0.610 \\
\(3\mathrm{o}3\mathrm{i}\) & 62 & 0.668 & 0.922/0.625/0.723 & 0.839 & 0.729 \\
\(3\mathrm{o}6\mathrm{i}\) & 64 & 0.716 & 0.937/0.616/0.721 & 0.877 & 0.741 \\
\(3\mathrm{o}12\mathrm{i}\) & 66 & 0.719 & 0.922/0.632/0.726 & 0.845 & 0.620 \\
\(3\mathrm{o}24\mathrm{i}\) & 58 & 0.749 & 0.919/0.634/0.730 & 0.843 & 0.609 \\
\bottomrule
\end{tabular}
\caption{GPT-5-mini scaling on 4-node graphs. Rows are grouped into pure observation, pure intervention, and mixed intervention blocks. Task accuracy favors pure observation, while graph-faithfulness favors mixed settings.}
\label{tab:scaling_mini_4nodes_metrics}
\end{table}

\begin{table}[!ht]
\centering
\small
\setlength{\tabcolsep}{4pt}
\begin{tabular}{lccccc}
\toprule
Setting & Acc. & Root \(F_1\) & All-edge P/R/\(F_1\) & Freq-edge \(F_1\) & Freq-weight \(F_1\) \\
\midrule
\multicolumn{6}{l}{\textit{Pure observation}} \\
\(5\mathrm{o}0\mathrm{i}\) & 30 & 0.214 & 0.853/0.168/0.265 & 0.622 & 0.310 \\
\(10\mathrm{o}0\mathrm{i}\) & 20 & 0.245 & 0.848/0.165/0.260 & 0.580 & 0.200 \\
\(20\mathrm{o}0\mathrm{i}\) & 14 & 0.271 & 0.812/0.174/0.272 & 0.584 & 0.187 \\
\midrule
\multicolumn{6}{l}{\textit{Pure intervention}} \\
\(0\mathrm{o}5\mathrm{i}\) & 12 & 0.095 & 0.298/0.114/0.154 & 0.206 & 0.179 \\
\(0\mathrm{o}10\mathrm{i}\) & 12 & 0.047 & 0.187/0.081/0.103 & 0.142 & 0.110 \\
\(0\mathrm{o}20\mathrm{i}\) & 16 & 0.073 & 0.243/0.108/0.145 & 0.177 & 0.137 \\
\midrule
\multicolumn{6}{l}{\textit{Mixed intervention}} \\
\(5\mathrm{o}0\mathrm{i}\) & 30 & 0.214 & 0.853/0.168/0.265 & 0.622 & 0.310 \\
\(5\mathrm{o}5\mathrm{i}\) & 26 & 0.378 & 0.868/0.344/0.434 & 0.661 & 0.505 \\
\(5\mathrm{o}10\mathrm{i}\) & 16 & 0.318 & 0.849/0.275/0.376 & 0.590 & 0.409 \\
\(5\mathrm{o}20\mathrm{i}\) & 26 & 0.342 & 0.841/0.320/0.428 & 0.676 & 0.438 \\
\bottomrule
\end{tabular}
\caption{GPT-5-mini scaling on 6-node graphs. Rows are grouped into pure observation, pure intervention, and mixed intervention blocks. Mixed settings improve graph recovery even when all end-task accuracies remain low.}
\label{tab:scaling_mini_6nodes_metrics}
\end{table}

Figure~\ref{fig:app_scaling_curves} then plots the same task-prediction axis
for both model families and graph sizes, making clear that better endpoint
frequency prediction does not by itself certify full causal recovery.

The stronger-model tables below expose the complementary failure mode. Even
when \texttt{GPT-5.2-high} solves most endpoint predictions under
observation-only budgets, the graph metrics still require mixed interaction.

\begin{table}[!ht]
\centering
\small
\setlength{\tabcolsep}{4pt}
\begin{tabular}{lccccc}
\toprule
Setting & Acc. & Root \(F_1\) & All-edge P/R/\(F_1\) & Freq-edge \(F_1\) & Freq-weight \(F_1\) \\
\midrule
\multicolumn{6}{l}{\textit{Pure observation}} \\
\(3\mathrm{o}0\mathrm{i}\) & 100 & 0.544 & 1.000/0.488/0.648 & 1.000 & 1.000 \\
\(6\mathrm{o}0\mathrm{i}\) & 100 & 0.557 & 1.000/0.488/0.648 & 1.000 & 1.000 \\
\(12\mathrm{o}0\mathrm{i}\) & 100 & 0.541 & 1.000/0.488/0.648 & 1.000 & 1.000 \\
\midrule
\multicolumn{6}{l}{\textit{Pure intervention}} \\
\(0\mathrm{o}3\mathrm{i}\) & 100 & 0.330 & 0.310/0.325/0.317 & 0.323 & 0.336 \\
\(0\mathrm{o}6\mathrm{i}\) & 100 & 0.320 & 0.314/0.315/0.314 & 0.330 & 0.348 \\
\(0\mathrm{o}12\mathrm{i}\) & 98 & 0.200 & 0.222/0.223/0.222 & 0.240 & 0.261 \\
\midrule
\multicolumn{6}{l}{\textit{Mixed intervention}} \\
\(3\mathrm{o}0\mathrm{i}\) & 100 & 0.544 & 1.000/0.488/0.648 & 1.000 & 1.000 \\
\(3\mathrm{o}3\mathrm{i}\) & 96 & 0.947 & 0.971/0.954/0.958 & 0.990 & 0.970 \\
\(3\mathrm{o}6\mathrm{i}\) & 96 & 0.947 & 0.980/0.935/0.949 & 0.987 & 0.987 \\
\(3\mathrm{o}12\mathrm{i}\) & 96 & 0.900 & 0.974/0.956/0.960 & 0.992 & 0.992 \\
\bottomrule
\end{tabular}
\caption{GPT-5.2-high scaling on 4-node graphs. Rows are grouped into pure observation, pure intervention, and mixed intervention blocks. Observation-only settings already solve the reactor task, while mixed settings best recover the full explicit graph.}
\label{tab:scaling_gpt52_4nodes_metrics}
\end{table}

\begin{table}[!ht]
\centering
\small
\setlength{\tabcolsep}{4pt}
\begin{tabular}{lccccc}
\toprule
Setting & Acc. & Root \(F_1\) & All-edge P/R/\(F_1\) & Freq-edge \(F_1\) & Freq-weight \(F_1\) \\
\midrule
\multicolumn{6}{l}{\textit{Pure observation}} \\
\(5\mathrm{o}0\mathrm{i}\) & 90 & 0.309 & 0.990/0.316/0.469 & 1.000 & 1.000 \\
\(10\mathrm{o}0\mathrm{i}\) & 86 & 0.313 & 1.000/0.316/0.471 & 1.000 & 0.980 \\
\(20\mathrm{o}0\mathrm{i}\) & 92 & 0.329 & 1.000/0.316/0.471 & 1.000 & 1.000 \\
\midrule
\multicolumn{6}{l}{\textit{Pure intervention}} \\
\(0\mathrm{o}5\mathrm{i}\) & 68 & 0.293 & 0.378/0.278/0.305 & 0.386 & 0.401 \\
\(0\mathrm{o}10\mathrm{i}\) & 76 & 0.130 & 0.278/0.254/0.259 & 0.318 & 0.327 \\
\(0\mathrm{o}20\mathrm{i}\) & 82 & 0.317 & 0.333/0.304/0.311 & 0.379 & 0.385 \\
\midrule
\multicolumn{6}{l}{\textit{Mixed intervention}} \\
\(5\mathrm{o}0\mathrm{i}\) & 90 & 0.309 & 0.990/0.316/0.469 & 1.000 & 1.000 \\
\(5\mathrm{o}5\mathrm{i}\) & 74 & 0.731 & 0.882/0.763/0.793 & 0.957 & 0.957 \\
\(5\mathrm{o}10\mathrm{i}\) & 72 & 0.687 & 0.874/0.778/0.793 & 0.939 & 0.920 \\
\(5\mathrm{o}20\mathrm{i}\) & 80 & 0.767 & 0.889/0.778/0.799 & 0.957 & 0.959 \\
\bottomrule
\end{tabular}
\caption{GPT-5.2-high scaling on 6-node graphs. Rows are grouped into pure observation, pure intervention, and mixed intervention blocks. Observation-only settings favor end-task success, while mixed settings favor explicit graph recovery.}
\label{tab:scaling_gpt52_6nodes_metrics}
\end{table}

Finally, Figure~\ref{fig:app_reeval_scaling} plots all-edge recovery directly.
The figure is placed after the numeric tables so the reader can first inspect
the per-suite values and then compare the aggregate trend across both models.

\subsection{Early-Commitment Diagnostics}
\label{app:early_commitment}

\begin{figure}[!htbp]
\centering
\setlength{\abovecaptionskip}{3pt}
\setlength{\belowcaptionskip}{-2pt}
\begin{minipage}[t]{0.64\linewidth}
\centering
\includegraphics[width=\linewidth]{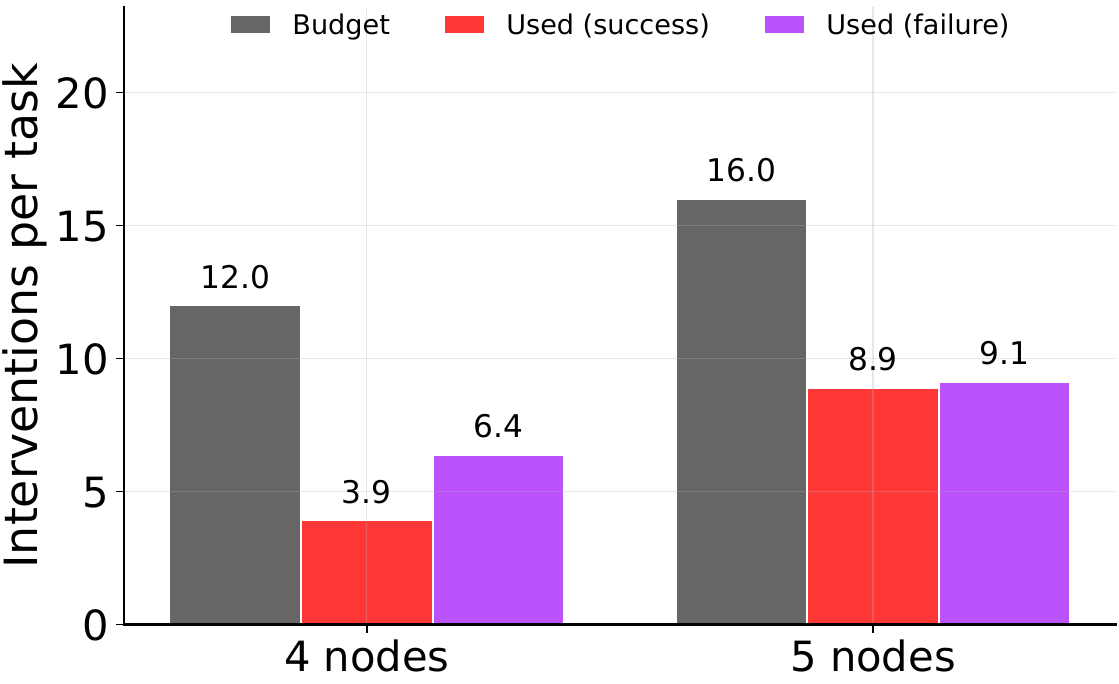}\\[-0.4em]
{\small (a) Budget vs.\ used.}
\end{minipage}

\vspace{-0.3em}
\begin{minipage}[t]{0.44\linewidth}
\centering
\includegraphics[width=\linewidth]{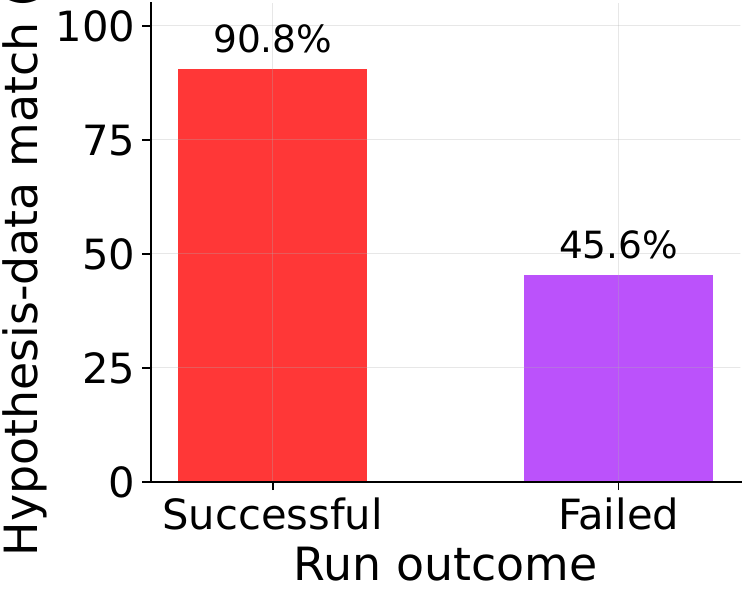}\\[-0.4em]
{\small (b) Hypothesis--data match.}
\end{minipage}\hfill
\begin{minipage}[t]{0.44\linewidth}
\centering
\includegraphics[width=\linewidth]{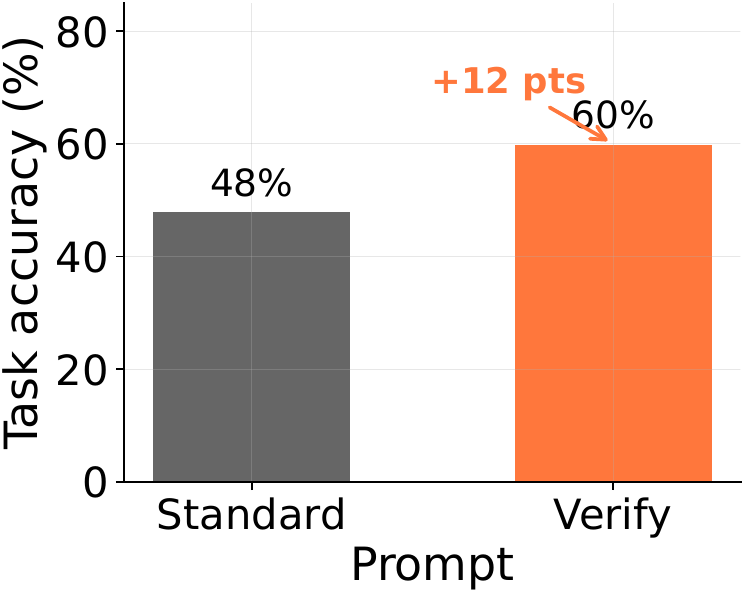}\\[-0.4em]
{\small (c) Verification gain.}
\end{minipage}
\caption{Early-commitment diagnostics for \texttt{GPT-5-mini}: both outcome
groups leave roughly half the budget unused; failed hypotheses fit collected
data worse; one verification step raises 4-node accuracy from 48\% to 60\%.}
\label{fig:overconfidence_diagnostics}
\end{figure}

Figure~\ref{fig:overconfidence_diagnostics} provides the full diagnostics for
the RQ4 early-commitment analysis in \S\ref{sec:rq4-failure}.

\end{document}